\documentclass[10pt,twocolumn,letterpaper]{article}
\usepackage[pagebackref,breaklinks,colorlinks]{hyperref}

\usepackage[pagenumbers]{cvpr} 

\usepackage{graphicx}
\usepackage{amsmath}
\usepackage{amssymb}
\usepackage{booktabs}
\usepackage{graphicx}
\usepackage[normalem]{ulem}
\usepackage{amsmath, amssymb}
\usepackage{footnotebackref}
\usepackage[para]{footmisc}
\usepackage{color}
\usepackage{enumerate}
\usepackage{graphbox}
\usepackage{makecell}
\usepackage{soul}
\usepackage{wrapfig}
\usepackage{graphics}
\usepackage{subcaption}
\usepackage{multirow}
\usepackage{eqparbox,array}
\usepackage{enumerate}
\usepackage{booktabs}
\usepackage{isotope}
\usepackage{siunitx}
\usepackage{tablists}
\usepackage{epsfig}
\usepackage{bm}
\usepackage[table]{ xcolor}
\usepackage{enumerate}
\usepackage{authblk}
\usepackage{enumitem}
\usepackage[pagebackref,breaklinks,colorlinks]{hyperref}
\makeatletter
\renewcommand\AB@affilsepx{, \protect\Affilfont}
\makeatother
\usepackage{lipsum}
\newcommand\blfootnote[1]{%
\begingroup
\renewcommand\thefootnote{}\footnote{#1}%
\addtocounter{footnote}{-1}%
\endgroup
}

\usepackage[capitalize]{cleveref}
\crefname{section}{Sec.}{Secs.}
\Crefname{section}{Section}{Sections}
\Crefname{table}{Table}{Tables}
\crefname{table}{Tab.}{Tabs.}

\def\cvprPaperID{6786} 
\def\confName{CVPR}
\def\confYear{2022}

\definecolor{darkgreen}{rgb}{0,0.6,0.2}
\definecolor{danred}{rgb}{0.9098,0.9098,0.9098}
\definecolor{shenred}{rgb}{0.8117,0.8117,0.8117}

\begin{document}

\title{M5Product: Self-harmonized Contrastive Learning  for E-commercial Multi-modal Pretraining} 
\author[1$\dagger$]{Xiao Dong}
\author[2$\dagger$]{Xunlin Zhan}
\author[1]{Yangxin Wu}
\author[3]{Yunchao Wei}
\author[4]{Michael C. Kampffmeyer}
\author[5]{Xiao-Yong Wei}
\author[6]{Minlong Lu}
\author[5]{Yaowei Wang}
\author[2$\star$ ]{Xiaodan Liang}
\affil[1]{Sun Yat-sen University}
\affil[2]{Shenzhen Campus of Sun Yat-sen University}
\affil[3]{Beijing Jiaotong University}
\affil[4]{UiT The Arctic University of Norway}
\affil[5]{PengCheng Laboratory}
\affil[6]{Alibaba Group
\protect\\
\textit {\small  \{dongx55, zhanxlin, wuyx29\}@mail2.sysu.edu.cn, \{dx.icandoit,wychao1987,xdliang328\}@gmail.com, ymlml@zju.edu.cn, michael.c.kampffmeyer@uit.no, cswei@scu.edu.cn, wangyw@pcl.ac.cn}}

\maketitle

\begin{abstract}
Despite the potential of multi-modal pre-training to learn highly discriminative feature representations from complementary data modalities, current progress is being slowed by the lack of large-scale modality-diverse datasets. By leveraging the natural suitability of E-commerce, where different modalities capture complementary semantic information, we contribute a large-scale multi-modal pre-training dataset \textbf{M5}Product. The dataset comprises 5 modalities (image, text, table, video, and audio), covers over 6,000 categories and 5,000 attributes, and is 500$\times$ larger than the largest publicly available dataset with a similar number of modalities. Furthermore, \textbf{M5}Product contains incomplete modality pairs and noise while also having a long-tailed distribution, resembling most real-world problems.
We further propose \textbf{S}elf-harmonized \textbf{C}ontr\textbf{A}stive \textbf{LE}arning (\textbf{SCALE}), a novel pretraining framework that integrates the different modalities into a unified model through an adaptive feature fusion mechanism, where the importance of each modality is learned directly from the modality embeddings and impacts the inter-modality contrastive learning and masked tasks within a multi-modal transformer model.
We evaluate the current multi-modal pre-training state-of-the-art approaches and benchmark their ability to learn from unlabeled data when faced with the large number of modalities in the \textbf{M5}Product dataset. We conduct extensive experiments on four downstream tasks and demonstrate the superiority of our \textbf{SCALE} model, providing insights into the importance of dataset scale and diversity. Dataset and codes are available at \footnote{\url{https://xiaodongsuper.github.io/M5Product_dataset/}}.
\blfootnote{$\dagger$ Equal contribution. $\star$ Corresponding Author.}

\end{abstract} 

\section{Introduction}
\label{sec:intro}

\begin{figure}
\centering
\includegraphics[width=0.45\textwidth]{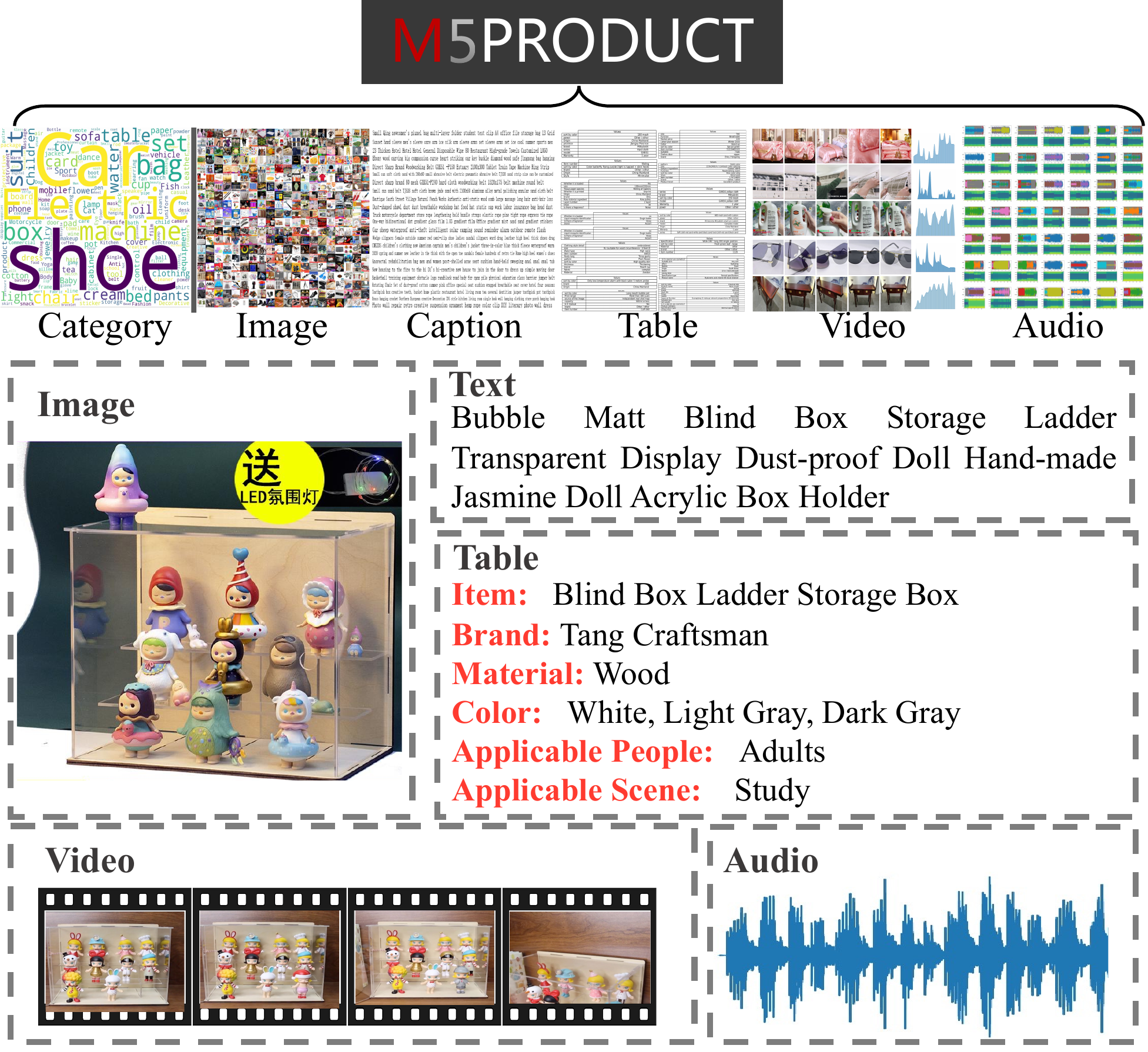} 
\vspace{-3mm}
\caption{Our \textbf{M}5Product dataset contains a large variety of modalities (image, text, table, video and audio) that depict the categories, descriptions, materials, properties and purposes of E-commerce products, and diverse real-world data samples.}
\label{fig:demo_pic}
\vspace{-5mm}
\end{figure}

Self-supervised learning has been driving the rapid development of fields such as computer vision and natural language processing, as well as research on multi-modal representation learning. In particular, it has been shown both from a theoretical~\cite{verify} and a practical~\cite{singlmodal1,singlmodal2} perspective that large scale datasets with \textbf{diverse modalities} can effectively enhance the discrimination of generated features and thus improve the performance in vision-language tasks. However, current advances are severely limited by the lack of such large-scale diverse-modality datasets, with the largest public multi-modal datasets only containing text and image modalities and no category information~\cite{CC}.

Given the prevalence of online shopping in daily life, with its natural occurrence of multi-modal information and diverse categories, multi-modal pre-training on E-commercial products has received increasing attention and led the developments of next-generation technology for several downstream tasks (e.g., multi-modal retrieval, multi-modal classification, and clustering). However, even among the present product datasets (e.g., RPC checkout~\cite{RPC}, Dress Retrieval~\cite{dress} and Product1M~\cite{Product1M}), the number of categories is insufficient to robustly verify the performance of downstream tasks.
More importantly, 
the current research community mostly focuses on two modalities (text and image) in both general multi-modal and E-commerce datasets, while ignoring the importance of additional complementary information from structural data as well as video and audio modalities.
Tabular data for instance can provide detailed information about properties and characteristics, such as brand, materials, attributes, and scenarios, while audio and video can convey different perspectives, scales, affordances, selling points, characteristics, and use scenarios that are not obvious from images or text alone. The focus on these two modalities is partly due to the lack of datasets with diverse modalities as well as an under-exploration of approaches to balance the modality importance in these settings. In particular, two key challenges are: 1) Modality Interaction: How to learn common representations from unimodal, bimodal, trimodal, and even multi-modal relationships between different modalities using an elegant approach that scales to a large number of modalities; 2) Modality Noise: How to reduce the influence of modality noise (missing and incompleted modalties) during the training process.


To address the problem of insufficient modality diversity and limited scale, while at the same time providing a challenging real-world scenario, we present a very large-scale E-commerce multi-modal product dataset \textbf{M5}Product, which is one of the largest and most diverse multi-modal product dataset to date. Our \textbf{M5}Product dataset contains more than 6 million multi-modal samples from 6,232 categories and has more complex and diverse modalities than existing datasets.  This allows \textbf{M5}Product to be used for more comprehensive evaluation of the practical application and generalization abilities of multi-modal pretraining models and can improve the modality fusion performance, facilitating new directions in multimodal research. Figure~\ref{fig:demo_pic} shows the five modalities (image, caption, video, audio, and specification (table)) of our dataset.

\begin{figure*}
\centering
\includegraphics[width=1\textwidth]{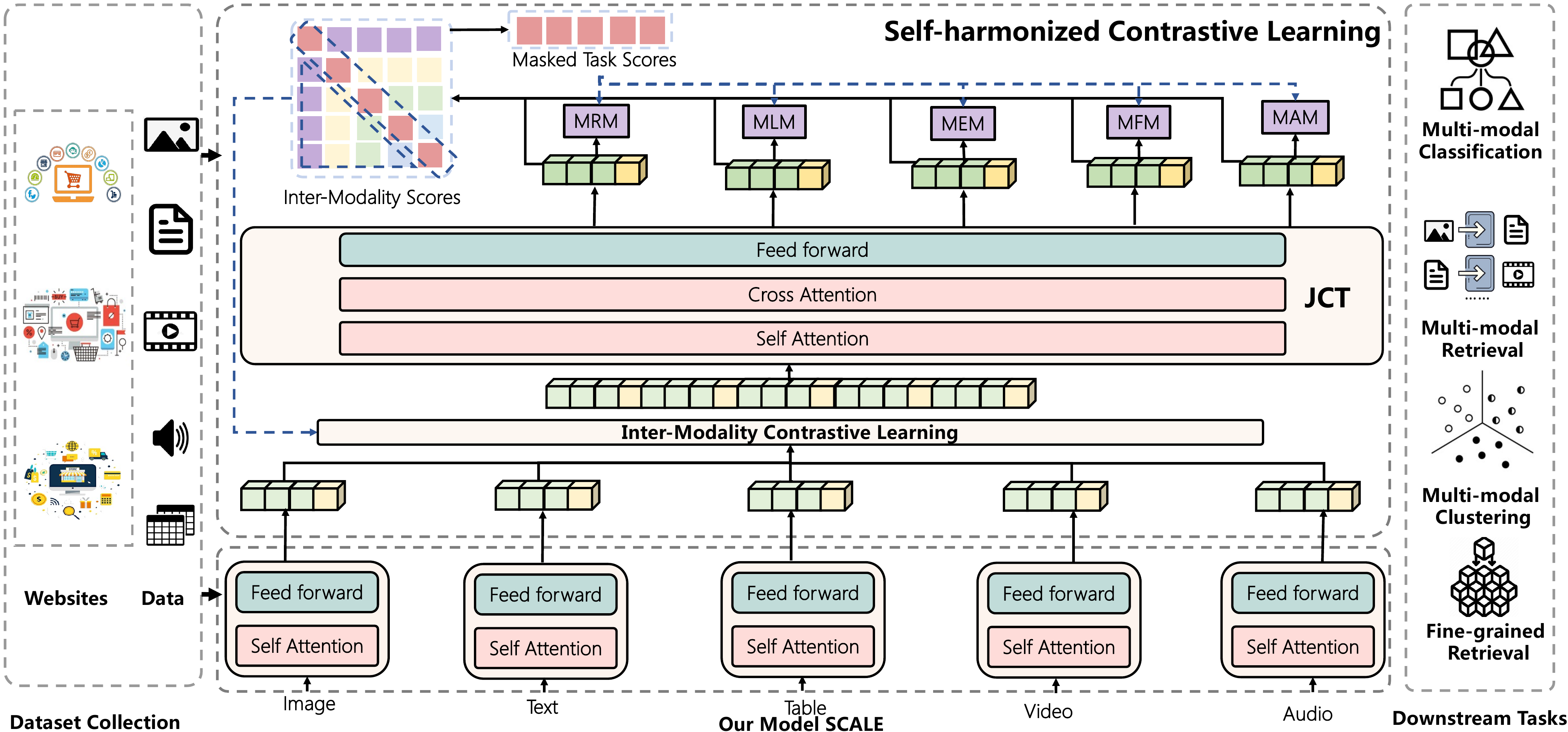} 
\vspace{-5mm}
\caption{An illustration of our \textbf{M5}Product benchmark. It consists of a five-modality E-commerce dataset with a more diverse and complex backgrounds collected from the real-world online-shopping website. It also proposes a \textbf{SCALE} model to capture the maximum modality complementary information for four common downstream tasks: 1) multi-modal retrieval, 2) fine-grained retrieval, 3) multi-modal classification, and 4) multi-modal clustering. The benchmark verifies the effectiveness of modality diversity in five widely used modalities.}
\vspace{-5mm}
\label{fig:benchmark}
\end{figure*}

To further address the modality fusion limitations of existing methods as well as handle modality noise, we propose a generic framework that takes five-modality data as inputs, as shown in Figure \ref{fig:benchmark}. The framework consists of a simple and efficient multi-modal five stream  pre-training model named \textbf{S}elf-harmonized \textbf{C}ontr\textbf{A}stive \textbf{LE}arning (\textbf{SCALE}) and is evaluated on several downstream tasks and compared with several recent state-of-the-art vision-language models~\cite{CLIP,ViLBERT,VLbert,VisualBERT,LXMERT,UNITER,CAPTURE}.
\textbf{SCALE} increases modality alignment effectiveness by implementing a self-harmonized strategy that adapts the alignment weights between different modalities in the contrastive learning modules and masked tasks to adaptively integrate complementary modality information.
In summary, our contributions are as follows:

\begin{itemize}[itemsep=0.5pt,topsep=1pt]
 \item  We provide the largest five-modality E-commerce dataset \textbf{M5Product}. Through its large scale, diversity, complex real scenarios and number of modalities, \textbf{M5Product} provides a comprehensive environment for evaluating the generalization performance of multi-modal pre-training models.
%
\item Our Self-harmonized Contrastive Learning (\textbf{SCALE}) framework learns adaptive modality interactions, resulting in more effective modality fusion. We compare  \textbf{SCALE} to a comprehensive set of baseline methods and demonstrate its superior performance on the M5Product dataset.
\item \emph{Interesting Observations}:  1) In large-scale and complex scenarios, the complementary gain of different modalities increases. Learning modality alignment weights allows our \textbf{SCALE} framework to effectively coordinate complementary information to achieve better performance. 2) For multi-modal pre-training models in the E-commerce domain, dataset scale and diversity are relatively important for the downstream tasks. Given the large-scale and diverse products, our \textbf{SCALE} framework generalizes better than other baselines to downstream tasks.
\end{itemize}

\section{Related Work}
\label{sec:related_work}

\noindent\textbf{Multi-modal pre-training datasets.}
Most multi-modal pre-training datasets are collected from social websites (e.g., Twitter and Facebook) and are limited to just two modalities collected for specified tasks.
These datasets can be divided into four categories according to their modality composition, i.e., audio/text, video/text, image/text, and others.
Among these, LJ Speech~\cite{ljspeech17} and SQuAD~\cite{SquAD} are classical audio/text datasets and used for voice synthesis and audio Q$\&$A, while most video/text datasets~\cite{TVQA,MovieQA,TGIF,AVSD,Youcook2,VATEX,MSRVTT,HowTo100M} are used for video Q$\&$A. However, these datasets commonly only contain a limited number of samples, limiting their applicability to multi-modal pretraining. Image/text datasets~\cite{CC,SBU,VG,COCO,Flickr,NLVR2,VQA,RPC,twitter100k,INRIA,nuswide,OpenImage}, on the other hand, tend to be larger and have been widely used for pretraining multi-modal models. Among these, the CC  3M~\cite{CC} with more than three million image-text pairs is the most widely used pre-training dataset, and has recently been expanded to CC 12M~\cite{CC12M}, the largest text-image cross-modal dataset currently. Apart from these, commonly used Image/text datasets for multi-modal retrieval tasks are MS COCO~\cite{COCO}, Flickr30K ~\cite{Flickr}, INRIA-Websearch~\cite{INRIA} and NUS-WIDE~\cite{nuswide} with standard annotations.
Other datasets include CMU-MOSEI~\cite{cmumosei} and XMedia~\cite{XMedia}, where CMU-MOSEI mainly focuses on the emotional analysis and XMedia is utilized for cross-modal retrieval.

Aside from the abovementioned datasets, there exist several E-commerce datasets. The Dress Retrieval~\cite{dress}, RPC checkout~\cite{RPC} and Product1M~\cite{Product1M} are typical E-commerce multi-modal datasets. The Dress Retrieval dataset contains 20,200 samples from 50 clothing categories, RPC checkout offers 30,000 samples of small retail goods on simple backgrounds and Product1M provides 1.18 million samples from 458 cosmetics classes. Compared with these three datasets, our \textbf{M5}Product is not only larger in terms of categories and data scale, but also contains a more diverse set of modalities. A detailed comparison with other multi-modal pre-training datasets is provided in Table~\ref{tab:dataset_compare}.


\noindent\textbf{Multi-modal pre-training for E-commerce products.} Several vision-language pre-training models have been explored for visual-text multi-modal learning in recent years. They can coarsely be divided into two categories: 1) Single-stream models whose transformer layer operates collectively on the concatenation of the visual and text inputs, e.g, VL-bert~\cite{VLbert}, Image-BERT~\cite{ImageBert}, VideoBERT~\cite{VideoBert},  MMT~\cite{MMT}, HERO~\cite{HERO}, VisualBERT~\cite{VisualBERT} and UNITER~\cite{UNITER}. 2) Dual-stream models whose image and text inputs are not concatenated, such as ViLBERT~\cite{ViLBERT}, LXMERT~\cite{LXMERT}, CLIP~\cite{CLIP} and DALL-E~\cite{DALL-E}.

\begin{table}
\centering
\caption{Comparisons with other widely used multi-modal datasets. "-" means not mentioned.  Our \textbf{M}5Product is one of the largest  multi-modal datasets compared with existing datasets. Six modalities are separately denoted as: Image (I), Text (T), Video (V), Audio (A), Table (Tab) and 3D Image (3D).}
\vspace{-3mm}
\label{tab:dataset_compare}
\setlength{\tabcolsep}{0.4mm}
\resizebox{1\columnwidth}{!}{
\begin{tabular}{c|cccc|ccc|c}
\toprule[1pt]
{{Dataset}}& {\footnotesize{}{Samples}} &  {\footnotesize{}{Categories}} &  {\footnotesize{}{Instances}}
&  {\footnotesize{}{Modalities}}
&  {\footnotesize{}{Modal type}}  & {\footnotesize{}{Product}}
\tabularnewline
\midrule[1pt]
SQuAD~\cite{SquAD} & 37,111  & - & - & 2 &A/T &no  \tabularnewline
HowTo100M~\cite{HowTo100M} &1,220,000  & 12 & - & 2 &V/T &no  \tabularnewline
CC 3M~\cite{CC} &3,300,000  & - & - & 2 &I/T &no  \tabularnewline
CC 12M~\cite{CC12M} & 12,423,374 & - & - &2 & I/T &no  \tabularnewline
CMU-MOSEI~\cite{cmumosei} &23,500  & 2 & - & 3 &T/V/A &no \tabularnewline
XMedia~\cite{XMedia} & 12,000  &20  & - &5 & I/T/V/A/3D &no \tabularnewline
\midrule[1pt]
RPC checkout~\cite{RPC} & 30,000 & 200 & 367,935 & 2 &I/T &yes  \tabularnewline
Dress Retrieval~\cite{dress} & 20,200 & 50 & $\sim$20,200 &2 & I/T &yes \tabularnewline
Product1M~\cite{Product1M} & 1,182,083  & 458 & 92,200 &2 & I/T &yes \tabularnewline
MEP-3M~\cite{MEP} & 3,012,959 & 599 & - &2 & I/T &yes  \tabularnewline
\rowcolor{danred}\textbf{M}5Product & \textbf{6,313,067}  & \textbf{6,232} & - &\textbf{5} & \textbf{I/T/V/A/Tab} &\textbf{yes} \tabularnewline
\bottomrule[1pt]
\end{tabular}}
\vspace{-5mm}
\end{table}

Within E-commerce, fashion-based tasks have been addressed in among others FashionBERT~\cite{Fashionbert}, MAAF~\cite{MAAF}, Kaleido-BERT~\cite{Kaleido-BERT}, M6~\cite{M6} and CAPTURE~\cite{Product1M}.
 All existing studies in the E-commerce scenarios focus solely on the image and text modalities and none of the approaches can utilize more modalities. 
Besides, all existing methods default to assigning the same contribution to different modalities when modeling multi-modal interactions.More specifically,  transformer-based approaches combine high-level features that are extracted from the different inputs via concatenation, where the uni-modal transformers are trained via masked task constraints or via constructing inter-modality losses between different modalities. This restricts the models from effectively prioritizing modalities and tends to limit performance improvements as the number of modalities increases. 

Our proposed benchmark fills this gap by exploiting all the diverse modalities of the \textbf{M5}Product dataset and provides a strong baseline for multi-modal pre-training research in the field of E-commerce and beyond.


\begin{figure}
\centering
\includegraphics[width=0.5\textwidth]{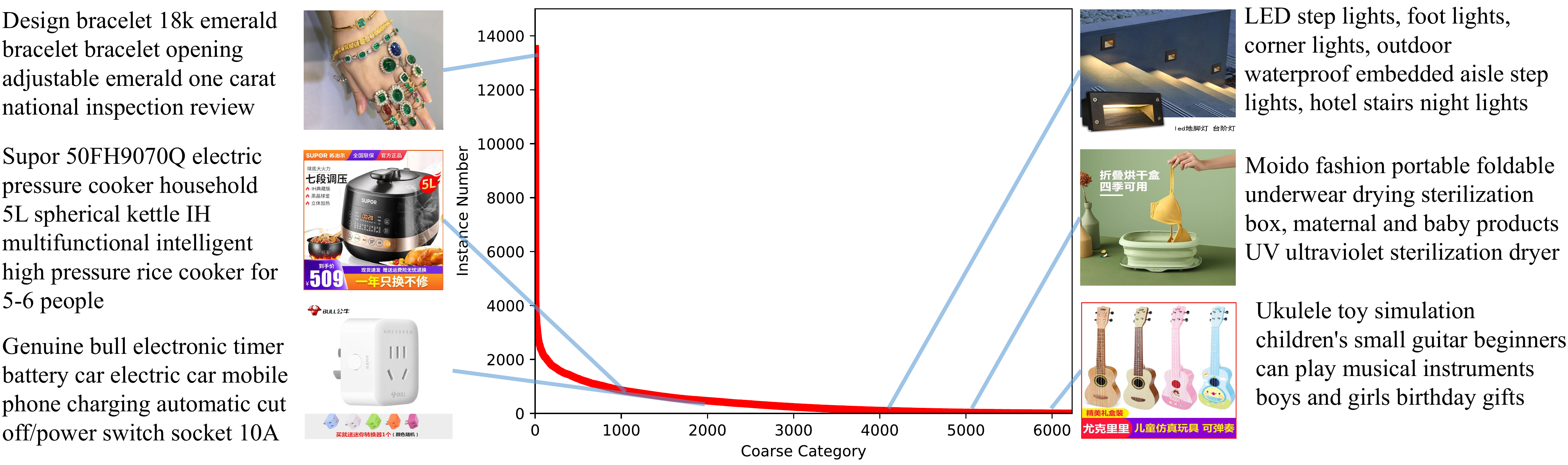}
\caption{Training data distribution over whole categories.}
\label{fig:data_distribution}
\vspace{-5mm}
\end{figure}



\section{M5Product Dataset}
\label{sec:text_product}

\noindent\textbf{Data Collections.}  The dataset is crawled from a popular E-commerce website~\footnote{We are authorised by the company to access and obtain the data. We are further authorised to share the dataset and the detailed license is given in Section~\ref{sec:license} of the supplementary material}. and the front page of each E-commerce product is analyzed to collect the multi-modal information consisting of product images, captions, videos, and specifications (table information)~\footnote{In this work we focus on core data modalities (image, text, video, audio, and table data) only and do not consider extracted feature representations such as OCR and Motion embeddings that are extracted from core modalities as separate modalities.}. Duplicate data is removed and audio information is extracted from videos via the \textbf{moviepy \footnote{https://pypi.org/project/moviepy/}} tool and saved in mp3 format. For product specifications, we extract 5,679 product properties and 24,398,673 values to construct a table database coarsely labeled by e-commerce merchants. After processing, the dataset contains 6,313,067 samples. Note, being a real-world dataset, our \textbf{M}5Product is, unlike traditional multi-modal datasets, not a complete paired dataset and contains samples with only a subset of modalities as well as long-tailed distributed (Figure~\ref{fig:data_distribution}).  We summarize the product characteristics that are relayed by the different modalities in our dataset in Table~\ref{tab:mainpoint}, where APP, USA, SPEC, SELL, PROD, MATE and CATE denote Appearance, Usage, Specification, Selling Point, Production, Material and Category Descriptions, respectively.

\begin{table}[]
\centering
\caption{The characteristics of different modalities for E-products.}
\vspace{-3mm}
\label{tab:mainpoint}
\setlength{\belowcaptionskip}{-0.3cm}
\large
\resizebox{0.95\columnwidth}{!}{
\begin{tabular}{c|ccccccc}
\toprule[1pt]
Modality &APP & USA & SPEC & SELL & PROD & MATE & CATE \\
\midrule[1pt]
Image & \checkmark& &&   &  & &\\
Text & &\checkmark & &\checkmark  & \checkmark &  &\checkmark  \\
Video &\checkmark &\checkmark &  &&\checkmark  &\checkmark  & \\
Audio &  &\checkmark  &  &  &\checkmark &\checkmark & \\
Table & &  &\checkmark & & \checkmark   & \checkmark & \checkmark \\
\toprule[1pt]
\end{tabular}
}
\vspace{-7mm}
\end{table}

\noindent \textbf{Quantitative analysis.} 1) \textbf{Diversity}:
The dataset consists of more than 6,000 classes covering various and massive amounts of E-commerce products such as clothes, cosmetics, and instruments. Figure~\ref{fig:demo_pic} illustrates the diversity of the modalities and categories and we further provide a description of the data format and the collection process in Section~\ref{sec:dataset_dataformat} of the supplementary materials. Finally, a quantitative analysis of the category and modality distributions can be found in Section~\ref{sec:dataset_dataanalysis}. Note that about 5$\%$ of products are unimodal samples \textit{e.g}. only either contain images, captions, or tabular properties. 2) \textbf{Quality}: We further provide a comparison between our \textbf{M}5Product dataset and some widely-used datasets for multi-modal pre-training in Table~\ref{tab:dataset_compare}. A more extensive comparison with other multi-modal datasets can be found in Section~\ref{sec:dataset_comp} of the supplementary materials. Compared with existing multi-modal datasets, \textbf{M}5Product is the first extremely large public real-world E-commerce product dataset that contains data of more than two modalities.


Moreover, our dataset contains a large amount of instances, i.e., more than six million samples from the 6,232 coarse categories. These abundant data will benefit several downstream tasks such as self-learning, weakly-supervised learning, multi-modal retrieval, cross-modal generation and fine-grained recognition.

\noindent \textbf{Additional analysis.} In the supplementary materials, we provide dataset collection details in Section~\ref{sec:dataset_collect} and detail how the dataset is split into training and test in Section~\ref{sec:dataset_split} and how annotations are obtained in Section~\ref{sec:dataset_anno}. We further provide a smaller split, referred to as \emph{subset}, which is used to show the difference in performance for a smaller dataset. Finally, we provide further insights into the composition of the dataset (missing modalities, unimodal data analysis, and data format) in supplementary  Section~\ref{sec:dataset_dataanalysis}.

\section{Our Methodology} 
\label{sec:experiment}

As shown in Figure~\ref{fig:benchmark}, our \textbf{SCALE} framework consists of a self-harmonized contrastive learning module and a self-supervised multi-modal transformer. In this section, we first provide the architectural design of \textbf{SCALE} in Section~\ref{subsec:SCL_archi} before describing the five masked tasks that enable the self-supervised learning of \textbf{SCALE} in Section~\ref{subsec:SCL_task}. Finally, we present the detailed learning process of \textbf{SCALE} and detail how multi-modal alignment is achieved in Section~\ref{subsec:SCL_SCL}.

\subsection{Architectural Design of \text{SCALE}}
\label{subsec:SCL_archi}

As depicted in Figure~\ref{fig:benchmark}, \textbf{SCALE} is a typical single-stream transformer architecture. In the bottom part, the Image/Text/Table/Video/Audio embedding layers and transformers aim to extract modality features and generate token features. Specifically, the text and table encoders are standard transformers to encode the caption and table information of products, respectively. The image encoder instead takes proposals extracted by bottom-up-attention~\cite{bottom_up} as inputs, while ordinal frames sampled from the video are fed into the video encoder. For the audio encoder, \textbf{SCALE} extracts MFCC\cite{2005Combining} features from audio. After being processed by the separate modality encoders, the token features of different modalities are concatenated and fed into a Joint Co-Transformer (\textbf{JCT}) module to capture the token relationships between different modalities.

\noindent\textbf{Missing Modalities.} Zero imputation of missing modalities is leveraged to utilize all available data when training \textbf{SCALE}. We provide experimental evidence that \textbf{SCALE} benefits from the incomplete samples in Section~\ref{sec:missing_data} of the supplementary material. 

\subsection{\textbf{SCALE} by Masked Multi-Modal Tasks}
\label{subsec:SCL_task}

Similar to previous works, we utilize several \textbf{pre}text tasks (\textbf{PRE}) to facilitate self-supervised learning of \textbf{SCALE} in the Joint Co-Transformer module. For modality-wise feature learning from the image and text modalities, we adopt the Masked Region Prediction task (MRP) and the Masked Language Modeling task (MLM), respectively, after the \textbf{JCT}. Utilizing the characteristics of the table, video, and audio modalities, we further propose a Mask Entity Modeling task (MEM), Mask Frame Prediction task (MFP), and Mask Audio Modeling task (MAM) following a similar strategy of predicting masked tokens. In all masked tasks, the ground-truth labels are the features of masked areas. For all masking tasks, 15$\%$ of the inputs are masked out and the remaining inputs are used to reconstruct the masked information. Please note that unlike in the MLM task, where 15$\%$ of individual words are masked, 15$\%$ of the entities (properties, brand names, etc.) are entirely masked out for the MEM task. This drives our model to learn better table representations to recover the masked inputs, which is illustrated in Section~\ref{subsec:aba_visual}. The loss function of the $i$th modality is defined as:
\begin{equation}
\label{eq:masked_task}
\mathcal{L}_{M_i}(\theta) = -E_{t_{msk} \sim \mathbf{t}} \log P_{\theta} \left( t_{msk} \mid t_{\neg msk}, \mathbf{M}_{\neg i} \right),
\end{equation}


\noindent where $t_{\neg msk}$ denotes the unmasked tokens surrounding the masked token $t_{msk}$, $\theta$ represents the network parameters, and $M_i$ and $M_{\neg i}$ are the $i$th modality and the remaining modalities, respectively.

\subsection{Self-harmonized Inter-Modality Contrastive Learning}
\label{subsec:SCL_SCL}
Self-harmonized Inter-Modality Contrastive Learning (\textbf{SIMCL}) is at the core of our proposed \textbf{SCALE} framework. It aims to facilitate the semantic alignment between different modalities via a self-harmonized strategy for adaptive \textbf{I}nter-\textbf{M}odality \textbf{C}ontrastive \textbf{L}earning (\textbf{IMCL}). For a minibatch of modality samples $D\in R^{B \times M \times F}$, where $B$, $M$ and $F$ denote the batch size, number of modalities, and embedding dimension, respectively, we first construct the contrastive loss between each modality.

Given $N$ data samples $\{(d_{i}^{(0)}$, $d_{i}^{(1)})\}_{i=1}^N$, where each sample has two modalities $(0)$ and $(1)$, we select the $N$ modality pairs as positive pairs in our contrastive learning. For each positive pair $(d_{i}^{(0)}$, $d_{i}^{(1)})$, negative pairs are constructed by pairing $d_{i}^{(0)}$ and $d_{i}^{(1)}$ with the remaining $N$-1 samples from the other modality, resulting in 2($N$-1) negative pairs. For a modality pair ($d_{i}^{(0)}$, $d_{i}^{(1)}$) and their embedding features ($f_{i}^{(0)}$, $f_{i}^{(1)}$), the cross-modal contrastive loss of each modality pair is:
\begin{equation}
    \label{eq:contrastive loss}
    \resizebox{.9\width}{!}{$\mathcal{L}_{CL}(d_{i}^{(0)}, d_i^{(1)}) = -\log \frac{\exp \left(\operatorname{sim}\left(f^{(0)}_i, f^{(1)}_i \right) /\tau \right)}{\sum\limits_{m=0}^{1}\sum\limits_{k=1}^{N}\bm{1}_{[k \neq i]}\exp \left(\operatorname{sim} \left(f_{i}^{(m)}, f_{k}^{(1-m)} \right) /\tau \right)}$},
\end{equation}

\noindent where $\operatorname{sim}$ is the cosine similarity, $\tau$ is the temperature parameter and $\bm{1}_{[k \neq i]}$ is a binary indicator function, and $\bm{1}$=1 for $k \neq i$ and 0 otherwise.

\begin{figure}
\centering
\includegraphics[width=0.47\textwidth]{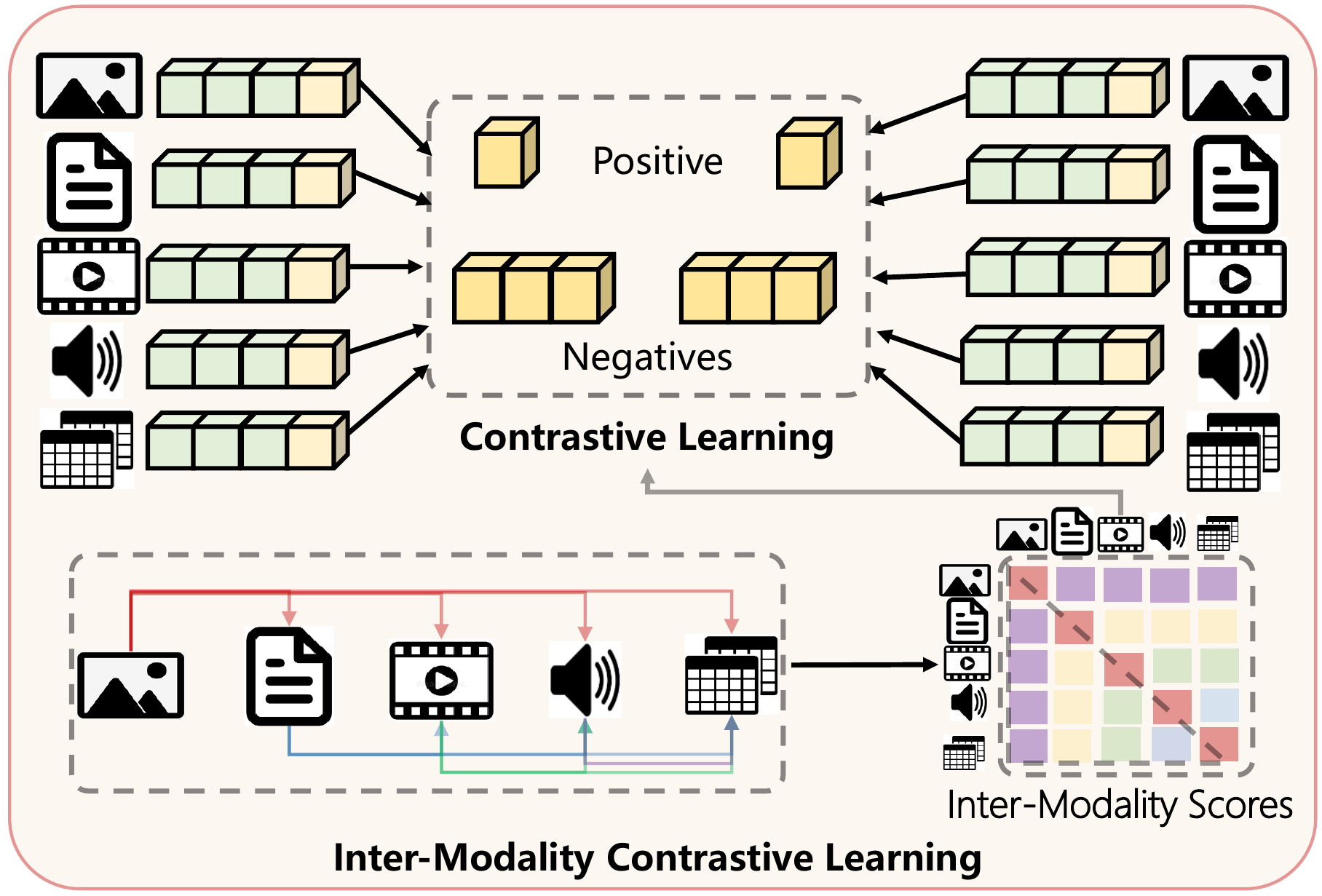}
\vspace{-3mm}
\caption{The Inter-Modality Contrastive Learning module of our \textbf{SCALE} framework.}
\vspace{-5mm}
\label{fig:IMCL}
\end{figure}


In most prior work, only two modalities are considered and Eq.~\ref{eq:contrastive loss} can be used. However, when considering trimodal data or data with even more than three modalities, it is not suitable to directly fit the loss function as it does not account for the difference in complementary information that different modalities contribute. To solve this problem, we define a simple but effective self-harmonized method to model the complementary process of the inter-modal relationships. We introduce a modality alignment score matrix, to encode the relationships among the inter-modal losses $\mathcal{L}_{CL}$ and the intra-modal losses $\mathcal{L}_{M_i}$.
The alignment score matrix $S$ for each data sample is initialized by a zero matrix and updated as free model parameters. To obtain modality importance scores for each modality combination, we apply the softmax function to $S$. Finally, the importance scores are multiplied to generate the modality alignment score $S$ as $S = S \cdot \operatorname{softmax}(S)$. The learning process is shown in Figure~\ref{fig:IMCL} and illustrates that \textbf{SIMCL} takes full advantage of the inter-modal relationships. Given the modality alignment score $S$, the triangular part $S_{\bigtriangledown}$ is selected to weight the inter-modal losses $\mathcal{L}_{CL}$ and the diagonal part $S_{\setminus}$ is utilized to constrain the intra-modal losses $\mathcal{L}_{M_i}$, resulting in the weighted loss:

\begin{equation}
    \label{eq:weight_loss}
    \resizebox{.9\width}{!}{
    $\mathcal{L}_{total} = \operatorname{\sum}_{S_{i,j}}^{S_{\bigtriangledown}} \mathcal{L}_{CL_{i,j}}(S_{i,j} logit_{i,j}) + \operatorname{\sum}_{S_{i}}^{S_{\setminus}} \mathcal{L}_{M_{i}}(S_{i} logit_{i})$}
\end{equation}

\noindent where logit is the the loss logit.

\begin{table*}[t!]
\centering
\caption{The (pretrain/finetune) performance gains from sequentially adding more modalities using \textbf{SCALE} on the subset (top) and the whole dataset (bottom). The retrieval performances are based on the features extracted from pretrain and finetune stages.}
\vspace{-3mm}
\label{tab:multidiverse_whole}
\small
\resizebox{1.8\columnwidth}{!}{
\begin{tabular}{
    c |
    c|
	c
	c
	c|
	c
	c
	c
}
\toprule[1pt]
{Modality}  & {Accuracy} &{mAP@1} & {mAP@5} & {mAP@10}  & {Prec@1} & {Prec@5} & {Prec@10}\\
\midrule
Text & 77.42   & 47.70 / 65.10  & 53.63 / 68.39  &51.59 / 66.99 &47.70 / 65.10 &30.96 / 44.89 &24.15 / 33.44\\
+Image & 79.58  &51.47 / 67.02 &56.16 / 69.85 &54.41 / 68.43 &51.47 / 67.02&33.41 / 46.29 &25.55 / 34.29 \\
+Table & 82.83  &57.14 / 67.97 &61.71 / 70.34 & 59.64 / 69.38 &57.14 / 67.97 &38.02 / 46.85 &28.99 / 34.36\\
+Video & 84.31  & 58.57 / 69.79 & 63.15 / 72.30 & 61.02 / 70.67 & 58.57 / 69.79 &39.26 / 47.44 & 29.56 / 34.78 \\
\rowcolor{danred}+Audio & \textbf{85.50} &\textbf{ 58.72 / 70.62}
&\textbf{63.17 / 73.02} &\textbf{61.05 / 71.50} &\textbf{58.72 / 70.62}  &\textbf{39.66 / 48.20}
&\textbf{30.32 / 35.35} \\
\midrule
Text &  81.11  & 55.82 / 69.47   & 60.74 / 72.74  & 59.02 / 71.79  & 55.82 / 69.47 & 36.99 / 48.76 & 28.04 / 35.84 \\
+Image & 83.68  & 59.81 / 71.51  & 64.13 / 74.51  & 62.18 / 73.21  & 59.81 / 71.51 & 38.97 / 49.27  & 30.15 / 36.72 \\
+Table &  84.63  & 61.32 / 72.34 &65.53 / 74.86  & 63.62 /
73.47  &61.32 / 72.34  & 40.66 / 49.77  & 30.78 / 36.95 \\
+Video & 84.90  & 62.65 / 72.59  & 65.67 / 75.05  & 63.87 / 73.62  & 62.65 / 72.59  & 41.18 / 49.96  & 31.01 / 37.04  \\
\rowcolor{danred}+Audio & \textbf{86.57} &\textbf{63.56  / 73.77}
&\textbf{67.51 / 76.17 } &\textbf{ 65.39 / 74.73 } &\textbf{ 63.56 / 74.01 }  &\textbf{ 42.68 / 50.78 }
&\textbf{ 32.17 / 37.42 } \\
\bottomrule[1pt]
\end{tabular}}
\vspace{-3mm}
\end{table*}

\begin{table*}[t!]
\centering
\caption{The performance of our model \textbf{SCALE} under different modality combinations on the coarse- and fine-grained multi-modal retrieval and classification tasks. In the following, I, T, Tab, V and A denote image, text, table, video and audio modalities, respectively.}
\vspace{-3mm}
\label{tab:scale_coarse_fine_perform}
\resizebox{1.8\columnwidth}{!}{
\begin{tabular}{
    c |
    c |
	c
	c
	c|
	c
	c
	c
}
\toprule[1pt]
Modality Combinations & Accuracy & {mAP@1} & {mAP@5} & {mAP@10} & {Prec@1} & {Prec@5} & {Prec@10}\\
\midrule
{I+Tab} & 62.00 & 44.53 / 45.97 & 49.62 / 51.89 & 48.28 / 50.33 & 44.53 / 45.97 & 30.89 / 34.08 & 23.65 / 28.63  \\
{I+V} & 34.57 & 20.57 / 36.29 & 26.78 / 42.72 & 26.41 / 41.38 & 20.57 / 36.29 & 14.71 / 26.52 & 11.78 / 22.34 \\
{I+A} & 27.67 & 15.73 / 35.64 & 20.85 / 42.96 & 20.72 / 41.70 & 15.73 / 35.64 & 11.16 / 27.02 & 9.47 / 22.78 \\
\rowcolor{danred}{\textbf{I+T}} & \textbf{79.58}  & \textbf{67.02 / 62.20} & \textbf{69.85 / 66.97} & \textbf{68.43 / 64.21}  & \textbf{67.02 / 62.20} & \textbf{46.29 / 49.85} & \textbf{34.29 / 42.36} \\
\midrule[1pt]
{I+T+V} & 80.34 & 67.35 / 63.05 & 70.29 / 67.37 & 68.95 / 64.62 & 67.35 / 63.05 & 46.45 / 50.85 & 34.33 / 43.02 \\
{I+T+A} & 79.73 & 67.19 / 64.21 & 70.15 / 68.25 & 68.64 / 65.35 & 67.19 / 64.21 & 46.33 / 50.42 & 33.32 / 42.93  \\
{I+Tab+V} & 63.09 & 45.94 / 47.33 & 51.32 / 53.33 & 49.78 / 51.28 & 45.94 / 47.33 & 31.69 / 35.81 &  24.12 / 30.05  \\
\rowcolor{danred}{\textbf{I+T+Tab}} & \textbf{82.83} & \textbf{67.97 / 68.30} & \textbf{70.34 / 72.67} & \textbf{69.38 / 70.07} & \textbf{67.97 / 68.30} & \textbf{46.85 / 57.44} & \textbf{34.36 / 50.59}  \\
\midrule[1pt]
{I+T+Tab+V}  & 84.31 & 69.79 / 68.40 & 72.30 / 72.91  & 70.67 / 70.31 & 69.79 / 68.40 & 47.44 / 57.60 & 34.78 / 51.47 \\
{I+Tab+A+V}  & 63.54 & 47.24 / 48.24 & 52.07 / 53.89 & 50.41 / 51.89 & 47.24 / 48.24  & 32.19 / 36.29 & 24.47 / 30.74 \\
{I+T+A+V}  & 80.36 & 68.80 / 66.43 & 70.84 / 71.12  & 69.71 / 68.16 & 68.80 / 66.43 & 47.24 / 54.03 & 34.57 / 47.53 \\
\rowcolor{danred}{\textbf{I+T+Tab+A}}  & \textbf{84.33} & \textbf{70.23 / 68.97} & \textbf{72.59 / 73.07}  & \textbf{70.94 / 70.77} & \textbf{70.23 / 68.97} & \textbf{47.58 / 57.89} & \textbf{35.33 / 51.60} \\
\midrule[1pt]
\rowcolor{shenred}{\textbf{I+T+Tab+A+V}}  & \textbf{85.50} & \textbf{70.62 / 69.25} & \textbf{73.02 / 74.08}  & \textbf{71.50 / 71.02} & \textbf{70.62 / 69.25} & \textbf{48.20 / 58.76} & \textbf{35.35 / 52.05} \\
\bottomrule[1pt]
\end{tabular}}
\vspace{-3mm}
\end{table*}


\begin{table}[t!]
\centering
\caption{Comparisons of image and text modalities on the subset (top) and the whole dataset (bottom).}
\vspace{-3mm}
\label{tab:class}
\resizebox{0.85\columnwidth}{!}{
\begin{tabular}{
    c |
    c |
	c|
	c
	c
	c
}
\toprule[1pt]
{Method} &{mAP@1}  &{Accuracy} & {NMI} & {Purity}\\
\midrule
Image$_{based}$ & 15.17 &  27.67 & 63.62 & 54.86 \\ 
BERT~\cite{Bert} & 47.70  & 77.42  & 76.35  & 68.80  \\
VL-BERT~\cite{VLbert} & 49.31  & 78.13  &  80.51 & 71.91  \\
ViLBERT~\cite{ViLBERT} & 49.18  &78.24  & 80.51 &  71.91     \\
VisualBERT~\cite{VisualBERT} & 49.20  & 78.41  & 81.23 & 72.39   \\
CLIP~\cite{CLIP} & 49.39 &  78.35   & 81.75 & 72.50   \\
UNITER~\cite{UNITER} & 49.87 &  78.54  & 82.71 & 73.58  \\
CAPTURE~\cite{CLIP} & 50.30 & 78.69 & 83.06 & 74.14  \\
\rowcolor{danred} \textbf{SCALE (Ours)} &  \textbf{51.47}& \textbf{ 79.58}  & \textbf{84.23} & \textbf{75.81}   \\
\midrule[1pt]
Image$_{based}$ & 22.67 & 30.14 & 67.49 & 59.64 \\ 
BERT~\cite{Bert} & 55.82  & 82.11  & 87.30 & 71.75  \\
CLIP~\cite{CLIP} & 57.73 & 82.60 & 90.49 & 76.48   \\
\rowcolor{danred} \textbf{SCALE (Ours)} & \textbf{59.81} & \textbf{83.68}  & \textbf{92.01} & \textbf{78.34}   \\
\bottomrule[1pt]
\end{tabular}}
\vspace{-3mm}
\end{table}

\section{Experiments}

\noindent\textbf{Implementation Details.}
We use BERT\cite{Bert}  to initialize the text transformer of our proposed \textbf{SCALE} framework, while the remaining transformers are randomly initialized. Both the single-modality encoders and the multi-modal fusion encoders consist of 6 transformer layers each, adding up to a total of 12 transformer layers. The hidden state size of each modality transformer is 768 and the maximum sequence length for the captions and tables are set to 36 and 64, respectively. Using the same setting as in~\cite{ViLBERT} \footnote{https://github.com/airsplay/py-bottom-up-attention}, we utilize Faster R-CNN~\cite{FRCNN} with a backbone ResNet101~\cite{Resnet} pre-trained on the Visual Genome dataset~\cite{VG} to extract region features of selected 10 to 36 bounding boxes with high-class detection probability for each image. We train \textbf{SCALE} with a total batch size of 64 for 5 epochs using the Adam optimizer~\cite{Adam} with a warm-up learning rate of 1$e$-4. Additional implementation details of our model are provided in Section~\ref{sec:implement} of the supplementary material.

\noindent\textbf{Baselines.}
We compare \textbf{SCALE} to the following eight alternative pre-training methods that utilize image and text modalities as well as combinations of both: Bert~\cite{Bert} (Text$_{based}$), Image$_{based}$, ViLBERT~\cite{ViLBERT}, CLIP~\cite{CLIP},  VL-BERT~\cite{VLbert}, VisualBERT~\cite{VisualBERT}, UNITER~\cite{UNITER} and CAPTURE~\cite{CAPTURE}.  Image$_{based}$ and BERT~\cite{Bert} are 12-layer transformers based on the MLM (Mask Language Modeling) or MRP (Mask Region Prediction) task using image or text modality, providing single-modal baselines for the product retrieval, classification, and clustering tasks.
To ensure a fair comparison, the same hidden size of 768 is chosen for all baselines.

\noindent\textbf{Evaluation.}
We consider the following four downstream tasks to evaluate the learned representations: 1) Multi-modal retrieval: This task aims to find the most relevant target products using combinations of two or more modalities. A pair is considered a match if both belong to the same category; 2) Fine-grained multi-modal retrieval: Retrieval on an instance level, where only samples of the same product (i.e. color, model, shape, and style) are considered a match \footnote{A more thorough definition of the term \emph{same products} and how instance-level labels are obtained is provided in the supplementary.};
3) Multi-modal classification: Product category classification given the multi-modal features extracted from the joint co-transformer of \textbf{SCALE} using a linear classifier;  and 4) Multi-modal clustering: Product category clustering using k-Means clustering and the same features as in the classification setting.
For product retrieval, we adopt the widely used metrics mean Average Precision (mAP) and Precision (Prec)~\cite{retrieval_method1,retrieval_method2,retrieval_method3} to evaluate the retrieval accuracy on the two retrieval tasks. For product classification and clustering, all methods are evaluated using the Classification Precision (Classification accuracy), Normalized Mutual Information (NMI)~\cite{NMI} and Purity metrics.
In all experiments, models are trained on the training split. The pre-trained model is then applied to extract the modality features of the gallery and test splits for the product retrieval and clustering tasks. For the classification task, we finetune the pre-trained model on the classification subset containing 1,805 categories/classes 
and utilize the finetuned model to extract the features of the classification test set.

\subsection{Modality Diversity}
\label{subsec:diversity}
To examine the performance of our proposed \textbf{SCALE} framework and to verify the benefits of diverse modalities and dataset scale, we train \textbf{SCALE} with an increasing number of modalities and observe the variations in classification and multi-modal retrieval performance both for the whole \textbf{M5}Product dataset and the subset. More specifically, fused features are extracted from the joint co-transformer (\textbf{JCT}) of our \textbf{SCALE} after finetuning for the classification task and after pre-training and finetuning for the (coarse) multi-modal retrieval task. Results in Table~\ref{tab:multidiverse_whole} %
show that performance increases across all settings as modalities are added, illustrating the benefit of complementary modality information to learning multi-modality feature representations. It can also be observed that modality gains are larger on the whole dataset, supporting \emph{Interesting Observation 1}.

We further provide results for an extensive set of modality combinations to verify \textbf{SCALE}s effectiveness in leveraging the diverse modalities of our \textbf{M5}Product dataset. Table~\ref{tab:scale_coarse_fine_perform} provides results for the coarse- and fine-grained multi-modal retrieval tasks as well as the classification task after finetuning the model. As in the previous experiment, noticeable improvements can be observed as additional modalities are added. In particular, the addition of the text modality leads to high modality gains, verifying the benefits of including more diverse modalities that can capture different views of the same product. Interestingly, performance on the coarse-grained retrieval task is significantly worse than on the fine-grained retrieval task in most cases, indicating the complexity of the \textbf{M5}Product dataset and the diversity of the products in each category.

\begin{table}[t!]
\centering
\caption{Ablation study of the \textbf{SIMCL} module.}
\vspace{-3mm}
\label{tab:scale_module_aba}
\small
\resizebox{1\columnwidth}{!}{
\begin{tabular}{
    c |
    c
    c ||
    c|
	c|
	c
}
\toprule[1pt]
{\#} &{IMCL}  &{PRE} & {Accuracy} &{mAP@1,5,10}  & {Prec@1,5,10}\\
\midrule
1 &  && 83.77 & 68.45 / 70.92 / 69.30 &67.56 / 46.37 / 34.12 \\
2 &\checkmark & &84.44 &69.14 / 71.96 / 70.13 &69.14 / 47.15 / 34.84 \\
3 & &\checkmark &84.09&69.31 / 71.59 / 69.85 &69.31 / 46.72 / 34.42 \\
\rowcolor{danred}4 &\checkmark &\checkmark &\textbf{85.50} &\textbf{70.62} / \textbf{73.02} / \textbf{71.50} &\textbf{70.62} / \textbf{48.20} / \textbf{35.35} \\
\bottomrule[1pt]
\end{tabular}}
\end{table}

\begin{table}[t!]
\centering
\caption{Analysis of different masked tasks (token mask (MLM) and entity mask (MEM)) for the table modality.
}
\vspace{-3mm}
\label{tab:scale_entity_aba}
\resizebox{0.9\columnwidth}{!}{
\begin{tabular}{
    c |
    c |
	c |
	c
}
\toprule[1pt]
\textbf{Tasks}  & Accuracy &{mAP@1,5,10}  & {Prec@1,5,10} \\
\midrule
{MLM} & 84.05 & 68.34 / 71.19 / 69.43 & 68.34 / 47.02 / 34.43 \\
\rowcolor{danred}{\textbf{MEM}} & \textbf{85.50} & \textbf{70.62} / \textbf{73.02} / \textbf{71.50} & \textbf{70.62} / \textbf{48.20} / \textbf{35.35}\\
\bottomrule[1pt]
\end{tabular}}
\end{table}

\begin{table}[t!]
\centering
\huge
\caption{Analysis of treating text and table modalities separately ({T/Tab}) or stacked together ({T+Tab}).}
\vspace{-3mm}
\label{tab:scale_text_aba}
\resizebox{0.9\columnwidth}{!}{
\begin{tabular}{
    c |
    c |
	c |
	c
}
\toprule[1pt]
\textbf{Formats} &{Accuracy} & {mAP@1,5,10} & {Prec@1,5,10} \\
\midrule
{T+Tab} & 84.61 & 70.15 / 72.19 / 70.49 & 69.15 / 47.40 / 34.40 \\
\rowcolor{danred} {T/Tab}  & \textbf{85.50} & \textbf{70.62} / \textbf{73.02}  / \textbf{71.50} & 7\textbf{0.62} / \textbf{48.20} / \textbf{35.35}  \\
\bottomrule[1pt]
\end{tabular}}
\end{table}

\begin{figure}[t!]
\centering
\hspace{10mm}
\begin{minipage}[t]{0.43\textwidth}
\centering
\includegraphics[width=1\textwidth]{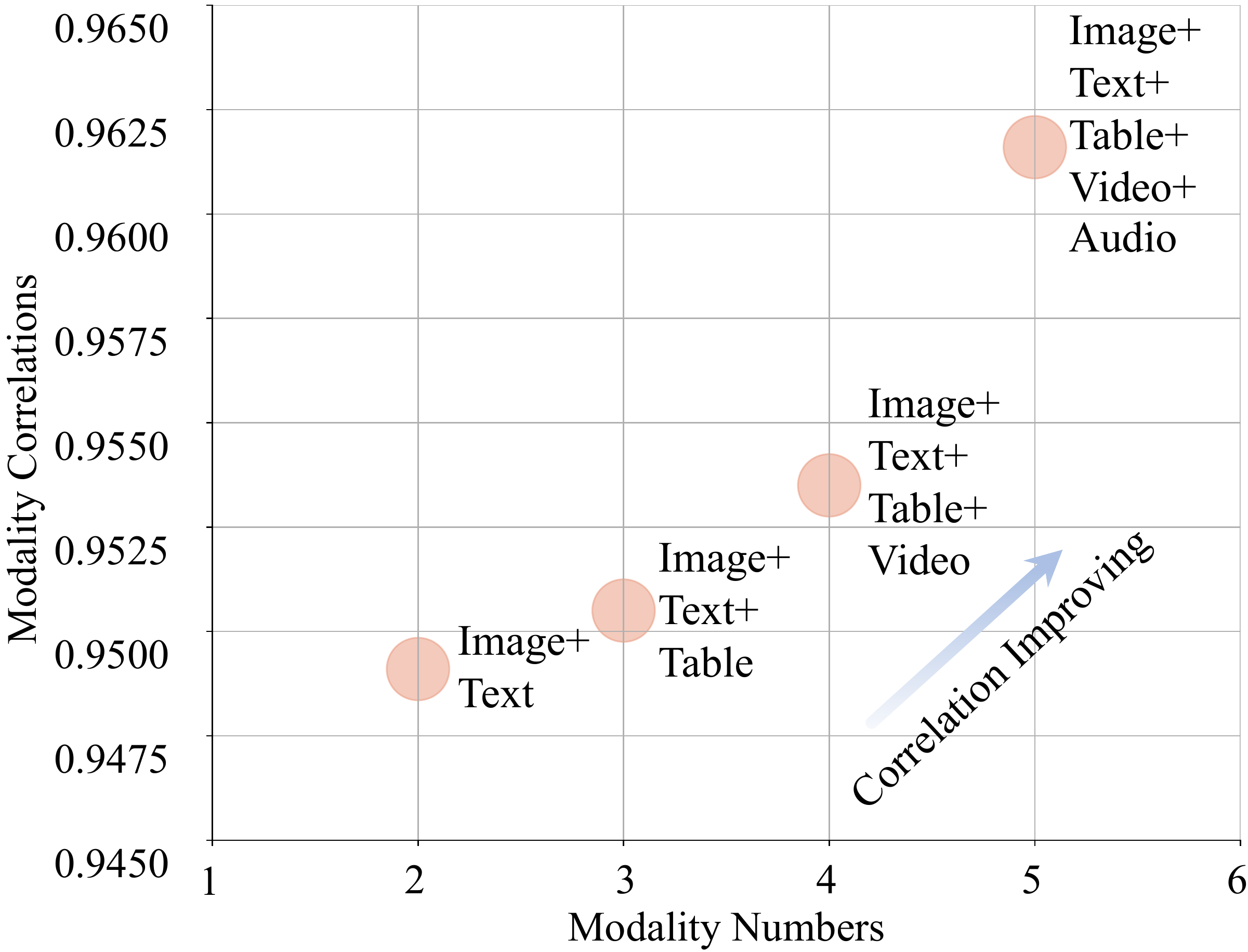}
\vspace{-4mm}
\caption{Variations of modality correlation gains with the number of modalities. 
}
\label{fig:correlation}
\end{minipage}
\vspace{-3mm}
\end{figure}



\noindent\textbf{Semantic Alignment.}
To additionally demonstrate the importance of modality diversity, we compute the modality correlation, the average cosine similarity between image and text features as obtained by the \textbf{JCT}, for an increasing number of modalities. Figure~\ref{fig:correlation} illustrates that the semantic alignment capability of the pre-training model increases as the number of modalities grows.

\subsection{Multi-modal Downstream Tasks}
\label{subsec:pre_task}


We evaluate \textbf{SCALE} on the \textbf{M5}Product dataset for the product retrieval, classification, and clustering tasks and compare results to several benchmark approaches in Table~\ref{tab:class}.
For the Image$_{based}$ and BERT~\cite{Bert} models, which only utilize the image and text features, respectively, the extracted features are fed directly into the classification model. For our \textbf{SCALE} approach, we utilize the fused modality features generated by the joint co-transformer, pre-trained on both image and text modalities. Only utilizing the image and text modalities allows us to facilitate a fair comparison to the recent state-of-the-art approaches ViLBERT~\cite{ViLBERT}, CLIP~\cite{CLIP},  VL-BERT~\cite{VLbert}, VisualBERT~\cite{VisualBERT}, UNITER~\cite{UNITER} and CAPTURE~\cite{CAPTURE}. Comparing our \textbf{SCALE} framework to the unimodal models, Image$_{based}$ and Bert~\cite{Bert}, we observe that exploiting multi-modal data significantly improves the performance across all tasks. We further observe that \textbf{SCALE}, by leveraging \textbf{SIMCL}, can efficiently fuse the modalities and outperform all the baseline approaches (\emph{Interesting Observation 2}).

\begin{figure}[t!]
\centering
\includegraphics[width=0.45\textwidth]{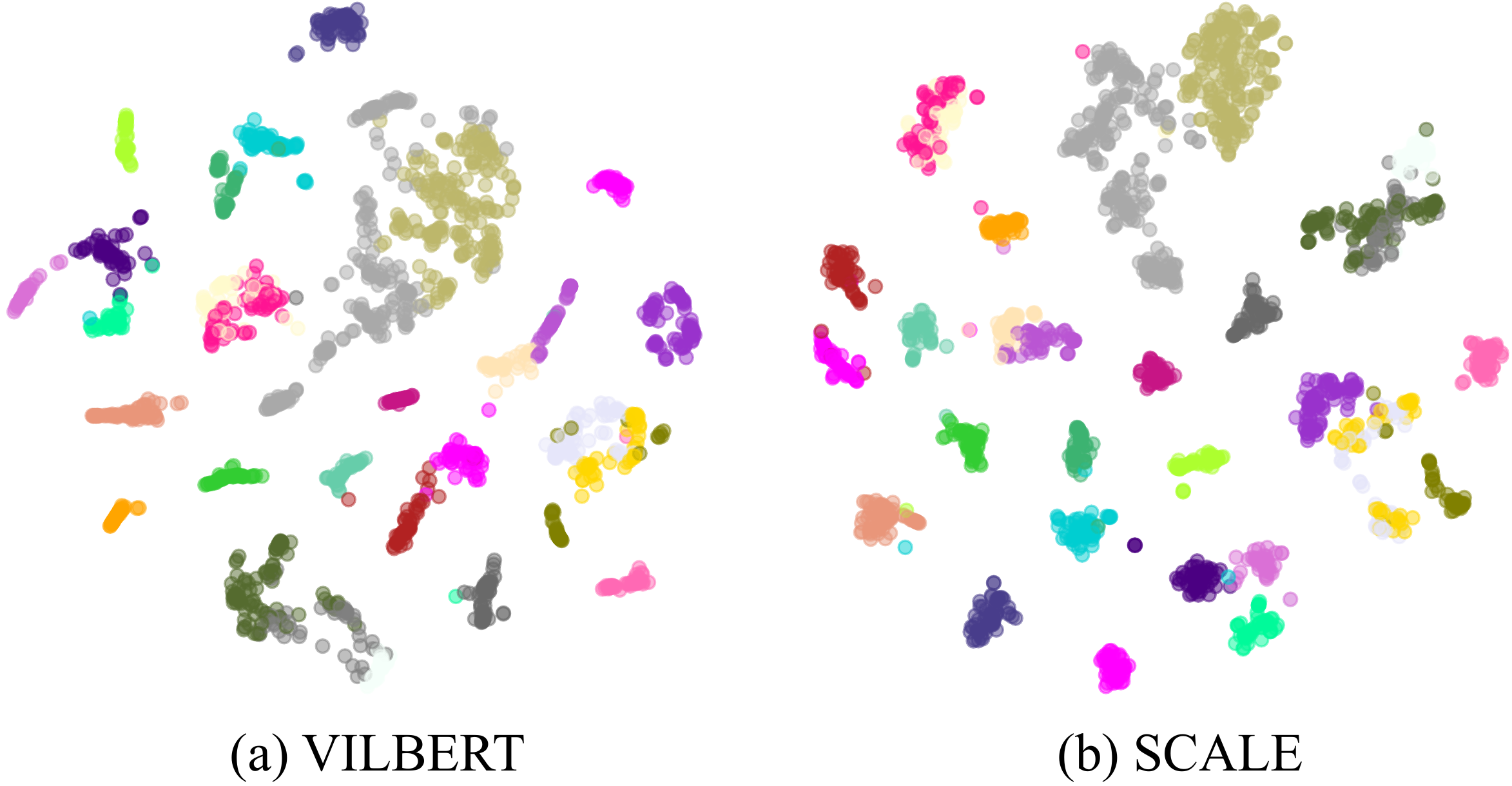}
\vspace{-3mm}
\caption{Visualize the embeddings generated by \textbf{SCALE} and
VILBERT via t-SNE. Points belonging to the same category are of
the same color. Best viewed in color.}
\label{fig:tsne_visual}
\vspace{-3.5mm}
\end{figure}

\begin{figure}[t!]
\centering
\begin{minipage}[t]{0.5\textwidth}
\centering
\includegraphics[width=1\textwidth]{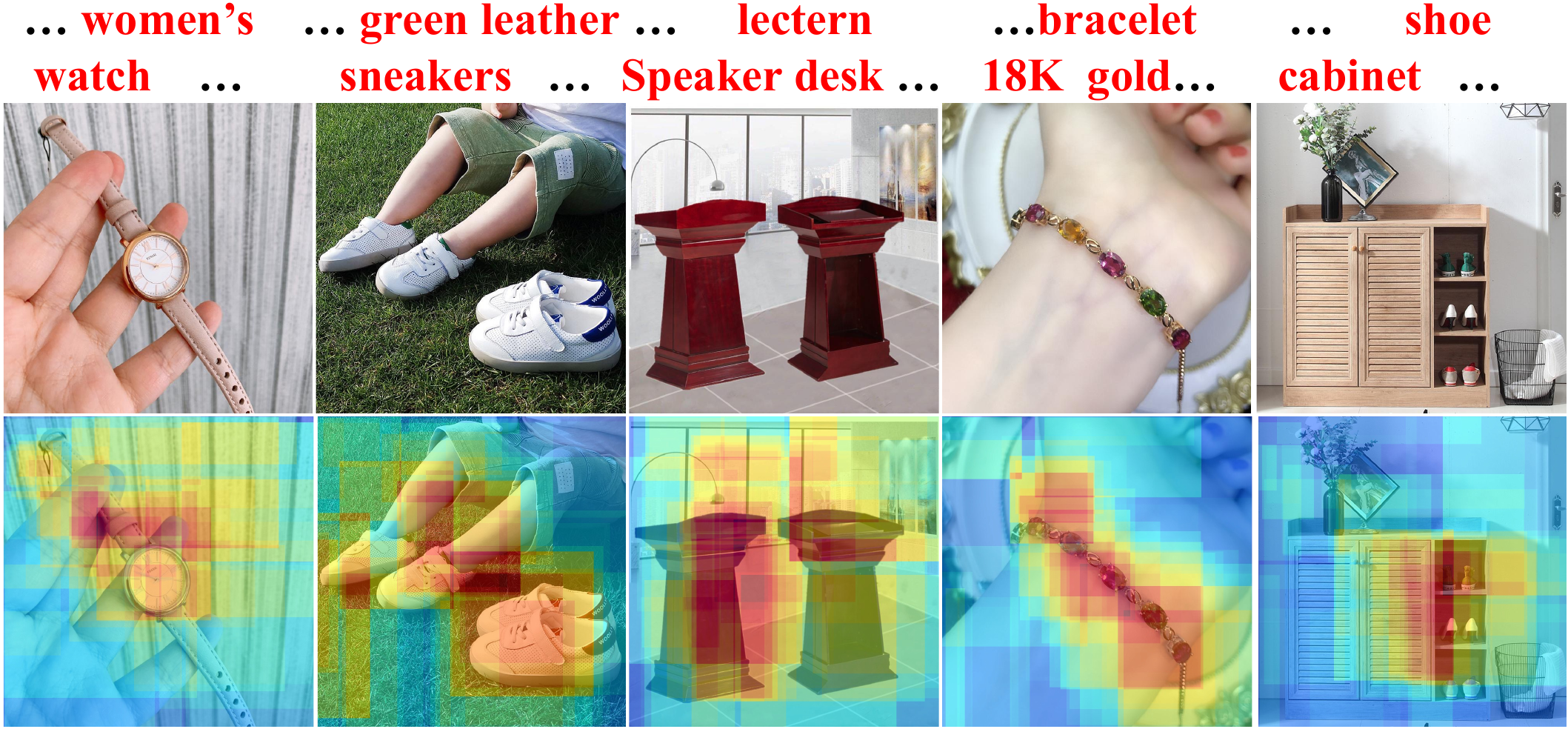}
\caption{
Attention attribution over proposals learned by our \textbf{SCALE}.}\vspace{-3.5mm}
\label{fig:atten_demo}
\end{minipage}
\end{figure}

\subsection{Ablation Studys and Visualization}
\label{subsec:aba_visual}
To explore how \textbf{SIMCL} influences \textbf{IMCL} and the \textbf{Pre}text tasks, we conduct several ablation studies. Table~\ref{tab:scale_module_aba} illustrates that improvements of approximately 2$\%$ are obtained in the classification task and more than 2$\%$ for the coarse-grained retrieval task when including both, highlighting the importance of both the \textbf{Pre}text tasks and the effective modality fusion of \textbf{SIMCL}.
We further analyze the effect of the MEM pretext task for the table modality and show the benefit of masking out complete entities over masking out individual tokens (MLM) in Table~\ref{tab:scale_entity_aba}. This benefit can be attributed to the fact that MEM ensures that \textbf{SCALE} learns representations that encode the semantic information of complete entities. Finally, we evaluate the performance of modelling the text and the table modalities using individual modality encoders and compare \textbf{SCALE}s retrieval performance to a baseline where text and table information is concatenated and fed to a single transformer, resembling the process of BERT~\cite{Bert}. By modelling both modalities individually, results in Table~\ref{tab:scale_text_aba} illustrate that more information can be preserved and we hypothesize that using a single transformer leads to a loss in table modality information for the benefit of the more expressive text modality.


Figure~\ref{fig:tsne_visual} shows t-SNE visualizations of the extracted features for the \textbf{JCT} module of our \textbf{SCALE} model and the alternative approach ViLBert~\cite{ViLBERT} on the \textbf{M5}Product dataset. \textbf{SCALE} not only better distinguishes different classes but also improves class compactness compared to the ViLBert model. Further, the attention attribution for different modalities are shown in Figure~\ref{fig:atten_demo} and verify that the visual features generated by \textbf{SCALE} are object-oriented and semantically interpretable.

\section{Limitations and future work} 
The experimental evaluation showed that \textbf{SCALE} is able to learn efficient representations from a large number of modalities for retrieval, classification, and clustering. However, more evaluation of the generative capabilities of the models representations is lacking and tasks such as image and caption generation could be promising directions to explore. We further provide some of \textbf{SCALE}s failure cases in supplementary Section~\ref{sec:failure}.

\noindent\textbf{Potential negative societal impact.} 
As a result of the strict ethical considerations used in the data collection process, where among others personally identifiable information has been removed, \textbf{M5}Product does not pose any ethical risks.


\section{Conclusion and Discussion}
\label{sec:conclusion}
To facilitate multi-modal pre-training, we present the \textbf{M5}Product dataset, which is the largest available multi-modal E-commerce product dataset, consisting of five core modalities (image, text, table, video, and audio). To further promote multi-modal research in retail and increase seller and buyer engagement, we also propose the novel \textbf{SCALE} multi-modal pre-training framework. By utilizing Self-harmonized Inter-Modality Contrastive Learning (\textbf{SIMCL}), \textbf{SCALE} is able to model and exploit modality relationships effectively and outperforms previous approaches on the \textbf{M5}Product multi-modal retrieval, classification, and clustering tasks. We believe that both the dataset and the proposed framework work will inspire research on scaling multi-modal pre-training beyond the commonly used image and text modalities.

\section{Acknowledgement}
This work was supported in part by Shenzhen Fundamental Research Program (Project No. RCYX20200714114642083, No. JCYJ20190807154211365),  National Key R$\&$D Program of China (2021ZD0112100), and CAAI-Huawei MindSpore Open Fund .  We thank MindSpore for the partial support of this work, which is a new deep learning computing framwork\footnote{https://www.mindspore.cn/}, and supported by Guangdong Provincial Key Laboratory of Fire Science and Intelligent Emergency Technology, Guangzhou 510006, China.

{\small
\bibliographystyle{ieee_fullname}
\bibliography{egbib}

\begin{thebibliography}{10}\itemsep=-1pt

\bibitem{OpenImage}
Open images dataset.
\newblock \url{https://storage.googleapis.com/openimages/web/index.html/},
  2018.

\bibitem{AVSD}
Huda AlAmri, Vincent Cartillier, Abhishek Das, Jue Wang, Anoop Cherian, Irfan
  Essa, Dhruv Batra, Tim~K. Marks, Chiori Hori, Peter Anderson, Stefan Lee, and
  Devi Parikh.
\newblock Audio visual scene-aware dialog.
\newblock In {\em CVPR}, pages 7558--7567, 2019.

\bibitem{bottom_up}
Peter Anderson, Xiaodong He, Chris Buehler, Damien Teney, Mark Johnson, Stephen
  Gould, and Lei Zhang.
\newblock Bottom-up and top-down attention for image captioning and visual
  question answering.
\newblock In {\em CVPR}, pages 6077--6086, 2018.

\bibitem{VQA}
Stanislaw Antol, Aishwarya Agrawal, Jiasen Lu, Margaret Mitchell, Dhruv Batra,
  C.~Lawrence Zitnick, and Devi Parikh.
\newblock {VQA}: {V}isual {Q}uestion {A}nswering.
\newblock In {\em ICCV}, pages 2425--2433, 2015.

\bibitem{CC12M}
Soravit Changpinyo, Piyush Sharma, Nan Ding, and Radu Soricut.
\newblock Conceptual 12m: Pushing web-scale image-text pre-training to
  recognize long-tail visual concepts.
\newblock In {\em CVPR}, pages 3558--3568, 2021.

\bibitem{MEP}
Delong Chen, Fan Liu, Xiaoyu Du, Ruizhuo Gao, and Feng Xu.
\newblock Mep-3m: A large-scale multi-modal e-commerce products dataset.
\newblock 2021.

\bibitem{UNITER}
Yen{-}Chun Chen, Linjie Li, Licheng Yu, Ahmed~El Kholy, Faisal Ahmed, Zhe Gan,
  Yu Cheng, and Jingjing Liu.
\newblock {UNITER:} universal image-text representation learning.
\newblock In Andrea Vedaldi, Horst Bischof, Thomas Brox, and Jan{-}Michael
  Frahm, editors, {\em ECCV}, volume 12375 of {\em Lecture Notes in Computer
  Science}, pages 104--120. Springer, 2020.

\bibitem{nuswide}
Tat{-}Seng Chua, Jinhui Tang, Richang Hong, Haojie Li, Zhiping Luo, and Yantao
  Zheng.
\newblock {NUS-WIDE:} a real-world web image database from national university
  of singapore.
\newblock In {\em CIVR}, 2009.

\bibitem{dress}
Charles Corbiere, Hedi Ben-Younes, Alexandre Ram{\'e}, and Charles Ollion.
\newblock Leveraging weakly annotated data for fashion image retrieval and
  label prediction.
\newblock In {\em ICCV Workshops}, pages 2268--2274, 2017.

\bibitem{Bert}
Jacob Devlin, Ming-Wei Chang, Kenton Lee, and Kristina Toutanova.
\newblock Bert: Pre-training of deep bidirectional transformers for language
  understanding.
\newblock {\em arXiv preprint arXiv:1810.04805}, 2018.

\bibitem{MAAF}
Eric Dodds, Jack Culpepper, Simao Herdade, Yang Zhang, and Kofi Boakye.
\newblock Modality-agnostic attention fusion for visual search with text
  feedback.
\newblock {\em arXiv preprint arXiv:2007.00145}, 2020.

\bibitem{MMT}
Valentin Gabeur, Chen Sun, Karteek Alahari, and Cordelia Schmid.
\newblock Multi-modal transformer for video retrieval.
\newblock In Andrea Vedaldi, Horst Bischof, Thomas Brox, and Jan{-}Michael
  Frahm, editors, {\em ECCV}, volume 12349, pages 214--229. Springer, 2020.

\bibitem{Fashionbert}
Dehong Gao, Linbo Jin, Ben Chen, Minghui Qiu, Peng Li, Yi Wei, Yi Hu, and Hao
  Wang.
\newblock Fashionbert: Text and image matching with adaptive loss for
  cross-modal retrieval.
\newblock In {\em SIGIR}, pages 2251--2260, 2020.

\bibitem{retrieval_method1}
Yunchao Gong, Svetlana Lazebnik, Albert Gordo, and Florent Perronnin.
\newblock Iterative quantization: {A} procrustean approach to learning binary
  codes for large-scale image retrieval.
\newblock {\em {IEEE} Trans. Pattern Anal. Mach. Intell.}, 35(12):2916--2929,
  2013.

\bibitem{Resnet}
Kaiming He, Xiangyu Zhang, Shaoqing Ren, and Jian Sun.
\newblock Deep residual learning for image recognition.
\newblock In {\em CVPR}, pages 770--778, 2016.

\bibitem{singlmodal1}
Jack Hessel and Lillian Lee.
\newblock Does my multimodal model learn cross-modal interactions? it's harder
  to tell than you might think!
\newblock In {\em EMNLP}, pages 861--877, 2020.

\bibitem{twitter100k}
Yuting Hu, Liang Zheng, Yi Yang, and Yongfeng Huang.
\newblock Twitter100k: {A} real-world dataset for weakly supervised cross-media
  retrieval.
\newblock {\em {IEEE} Trans. Multim.}, 20(4):927--938, 2018.

\bibitem{verify}
Yu Huang, Chenzhuang Du, Zihui Xue, Xuanyao Chen, Hang Zhao, and Longbo Huang.
\newblock What makes multimodal learning better than single (provably).
\newblock {\em arXiv preprint arXiv:2106.04538}, 2021.

\bibitem{ljspeech17}
Keith Ito and Linda Johnson.
\newblock The lj speech dataset.
\newblock \url{https://keithito.com/LJ-Speech-Dataset/}, 2017.

\bibitem{TGIF}
Yunseok Jang, Yale Song, Youngjae Yu, Youngjin Kim, and Gunhee Kim.
\newblock {TGIF-QA:} toward spatio-temporal reasoning in visual question
  answering.
\newblock In {\em CVPR}, pages 1359--1367, 2017.

\bibitem{Adam}
Diederik~P. Kingma and Jimmy Ba.
\newblock Adam: {A} method for stochastic optimization.
\newblock In {\em ICLR}, 2015.

\bibitem{INRIA}
Josip Krapac, Moray Allan, Jakob~J. Verbeek, and Fr{\'{e}}d{\'{e}}ric Jurie.
\newblock Improving web image search results using query-relative classifiers.
\newblock In {\em CVPR}, pages 1094--1101, 2010.

\bibitem{VG}
Ranjay Krishna, Yuke Zhu, Oliver Groth, Justin Johnson, Kenji Hata, Joshua
  Kravitz, Stephanie Chen, Yannis Kalantidis, Li{-}Jia Li, David~A. Shamma,
  Michael~S. Bernstein, and Li Fei{-}Fei.
\newblock Visual genome: Connecting language and vision using crowdsourced
  dense image annotations.
\newblock {\em Int. J. Comput. Vis.}, 123(1):32--73, 2017.

\bibitem{TVQA}
Jie Lei, Licheng Yu, Mohit Bansal, and Tamara~L. Berg.
\newblock {TVQA:} localized, compositional video question answering.
\newblock In {\em EMNLP}, pages 1369--1379, 2018.

\bibitem{SquAD}
Chia{-}Hsuan Li, Szu{-}Lin Wu, Chi{-}Liang Liu, and Hung{-}yi Lee.
\newblock Spoken squad: {A} study of mitigating the impact of speech
  recognition errors on listening comprehension.
\newblock In {\em ISCA}, pages 3459--3463, 2018.

\bibitem{HERO}
Linjie Li, Yen{-}Chun Chen, Yu Cheng, Zhe Gan, Licheng Yu, and Jingjing Liu.
\newblock {HERO:} hierarchical encoder for video+language omni-representation
  pre-training.
\newblock In {\em EMNLP}, pages 2046--2065, 2020.

\bibitem{VisualBERT}
Liunian~Harold Li, Mark Yatskar, Da Yin, Cho{-}Jui Hsieh, and Kai{-}Wei Chang.
\newblock Visualbert: {A} simple and performant baseline for vision and
  language.
\newblock {\em CoRR}, abs/1908.03557, 2019.

\bibitem{M6}
Junyang Lin, Rui Men, An Yang, Chang Zhou, Ming Ding, Yichang Zhang, Peng Wang,
  Ang Wang, Le Jiang, Xianyan Jia, Jie Zhang, Jianwei Zhang, Xu Zou, Zhikang
  Li, Xiaodong Deng, Jie Liu, Jinbao Xue, Huiling Zhou, Jianxin Ma, Jin Yu,
  Yong Li, Wei Lin, Jingren Zhou, Jie Tang, and Hongxia Yang.
\newblock {M6:} {A} chinese multimodal pretrainer.
\newblock {\em CoRR}, abs/2103.00823, 2021.

\bibitem{COCO}
Tsung{-}Yi Lin, Michael Maire, Serge~J. Belongie, James Hays, Pietro Perona,
  Deva Ramanan, Piotr Doll{\'{a}}r, and C.~Lawrence Zitnick.
\newblock Microsoft {COCO:} common objects in context.
\newblock In {\em ECCV}, pages 740--755, 2014.

\bibitem{ViLBERT}
Jiasen Lu, Dhruv Batra, Devi Parikh, and Stefan Lee.
\newblock Vilbert: Pretraining task-agnostic visiolinguistic representations
  for vision-and-language tasks.
\newblock In {\em NIPS}, pages 13--23, 2019.

\bibitem{retrieval_method3}
Xiaoqiang Lu, Xiangtao Zheng, and Xuelong Li.
\newblock Latent semantic minimal hashing for image retrieval.
\newblock {\em {IEEE} Trans. Image Process.}, 26(1):355--368, 2017.

\bibitem{HowTo100M}
Antoine Miech, Dimitri Zhukov, Jean{-}Baptiste Alayrac, Makarand Tapaswi, Ivan
  Laptev, and Josef Sivic.
\newblock Howto100m: Learning a text-video embedding by watching hundred
  million narrated video clips.
\newblock In {\em ICCV}, pages 2630--2640, 2019.

\bibitem{2005Combining}
Ksr Murty and B. Yegnanarayana.
\newblock Combining evidence from residual phase and mfcc features for speaker
  recognition.
\newblock {\em IEEE Signal Processing Letters}, 13(1):52--55, 2005.

\bibitem{SBU}
Vicente Ordonez, Girish Kulkarni, and Tamara~L. Berg.
\newblock Im2text: Describing images using 1 million captioned photographs.
\newblock In {\em NIPS}, pages 1143--1151, 2011.

\bibitem{pytorch}
Adam Paszke, Sam Gross, Soumith Chintala, Gregory Chanan, Edward Yang, Zachary
  DeVito, Zeming Lin, Alban Desmaison, Luca Antiga, and Adam Lerer.
\newblock Automatic differentiation in pytorch.
\newblock In {\em NIPS Workshop}, 2017.

\bibitem{XMedia}
Yuxin Peng, Xin Huang, and Yunzhen Zhao.
\newblock An overview of cross-media retrieval: Concepts, methodologies,
  benchmarks, and challenges.
\newblock {\em {IEEE} Trans. Circuits Syst. Video Technol.}, 28(9):2372--2385,
  2018.

\bibitem{ImageBert}
Di Qi, Lin Su, Jia Song, Edward Cui, Taroon Bharti, and Arun Sacheti.
\newblock Imagebert: Cross-modal pre-training with large-scale weak-supervised
  image-text data.
\newblock {\em arXiv preprint arXiv:2001.07966}, 2020.

\bibitem{CLIP}
Alec Radford, Jong~Wook Kim, Chris Hallacy, Aditya Ramesh, Gabriel Goh,
  Sandhini Agarwal, Girish Sastry, Amanda Askell, Pamela Mishkin, Jack Clark,
  Gretchen Krueger, and Ilya Sutskever.
\newblock Learning transferable visual models from natural language
  supervision.
\newblock In {\em ICML}, pages 8748--8763, 2021.

\bibitem{DALL-E}
Aditya Ramesh, Mikhail Pavlov, Gabriel Goh, Scott Gray, Chelsea Voss, Alec
  Radford, Mark Chen, and Ilya Sutskever.
\newblock Zero-shot text-to-image generation.
\newblock In Marina Meila and Tong Zhang, editors, {\em ICML}, volume 139 of
  {\em Proceedings of Machine Learning Research}, pages 8821--8831. {PMLR},
  2021.

\bibitem{FRCNN}
Shaoqing Ren, Kaiming He, Ross~B. Girshick, and Jian Sun.
\newblock Faster {R-CNN:} towards real-time object detection with region
  proposal networks.
\newblock {\em {IEEE} Trans. Pattern Anal. Mach. Intell.}, 39(6):1137--1149,
  2017.

\bibitem{CC}
Piyush Sharma, Nan Ding, Sebastian Goodman, and Radu Soricut.
\newblock Conceptual captions: {A} cleaned, hypernymed, image alt-text dataset
  for automatic image captioning.
\newblock In {\em ACL}, pages 2556--2565, 2018.

\bibitem{VLbert}
Weijie Su, Xizhou Zhu, Yue Cao, Bin Li, Lewei Lu, Furu Wei, and Jifeng Dai.
\newblock {VL-BERT:} pre-training of generic visual-linguistic representations.
\newblock In {\em ICLR}, 2020.

\bibitem{NLVR2}
Alane Suhr, Stephanie Zhou, Ally Zhang, Iris Zhang, Huajun Bai, and Yoav Artzi.
\newblock A corpus for reasoning about natural language grounded in
  photographs.
\newblock In {\em ACL}, pages 6418--6428, 2019.

\bibitem{VideoBert}
Chen Sun, Austin Myers, Carl Vondrick, Kevin Murphy, and Cordelia Schmid.
\newblock Videobert: {A} joint model for video and language representation
  learning.
\newblock In {\em ICCV}, pages 7463--7472, 2019.

\bibitem{LXMERT}
Hao Tan and Mohit Bansal.
\newblock {LXMERT:} learning cross-modality encoder representations from
  transformers.
\newblock In {\em EMNLP}, pages 5099--5110, 2019.

\bibitem{MovieQA}
Makarand Tapaswi, Yukun Zhu, Rainer Stiefelhagen, Antonio Torralba, Raquel
  Urtasun, and Sanja Fidler.
\newblock Movieqa: Understanding stories in movies through question-answering.
\newblock In {\em CVPR}, pages 4631--4640, 2016.

\bibitem{VATEX}
Xin Wang, Jiawei Wu, Junkun Chen, Lei Li, Yuan{-}Fang Wang, and William~Yang
  Wang.
\newblock Vatex: {A} large-scale, high-quality multilingual dataset for
  video-and-language research.
\newblock In {\em ICCV}, pages 4580--4590, 2019.

\bibitem{RPC}
Xiu-Shen Wei, Quan Cui, Lei Yang, Peng Wang, and Lingqiao Liu.
\newblock Rpc: A large-scale retail product checkout dataset.
\newblock {\em arXiv preprint arXiv:1901.07249}, 2019.

\bibitem{retrieval_method2}
Yair Weiss, Antonio Torralba, and Robert Fergus.
\newblock Spectral hashing.
\newblock In {\em NIPS}, pages 1753--1760, 2008.

\bibitem{GLAD}
Jacob Whitehill, Paul Ruvolo, Tingfan Wu, Jacob Bergsma, and Javier~R.
  Movellan.
\newblock Whose vote should count more: Optimal integration of labels from
  labelers of unknown expertise.
\newblock In {\em NIPS}, pages 2035--2043, 2009.

\bibitem{MSRVTT}
Jun Xu, Tao Mei, Ting Yao, and Yong Rui.
\newblock {MSR-VTT:} {A} large video description dataset for bridging video and
  language.
\newblock In {\em CVPR}, pages 5288--5296, 2016.

\bibitem{NMI}
Chengfu Yang and Zhang Yi.
\newblock Document clustering using locality preserving indexing and support
  vector machines.
\newblock {\em Soft Comput.}, 12(7):677--683, 2008.

\bibitem{Flickr}
Peter Young, Alice Lai, Micah Hodosh, and Julia Hockenmaier.
\newblock From image descriptions to visual denotations: New similarity metrics
  for semantic inference over event descriptions.
\newblock {\em Trans. Assoc. Comput. Linguistics}, 2:67--78, 2014.

\bibitem{cmumosei}
Amir Zadeh, Paul~Pu Liang, Soujanya Poria, Erik Cambria, and Louis{-}Philippe
  Morency.
\newblock Multimodal language analysis in the wild: {CMU-MOSEI} dataset and
  interpretable dynamic fusion graph.
\newblock In {\em ACL}, pages 2236--2246, 2018.

\bibitem{Product1M}
Xunlin Zhan, Yangxin Wu, Xiao Dong, Yunchao Wei, Minlong Lu, Yichi Zhang, Hang
  Xu, and Xiaodan Liang.
\newblock Product1m: Towards weakly supervised instance-level product retrieval
  via cross-modal pretraining.
\newblock In {\em ICCV}, 2021.

\bibitem{CAPTURE}
Xunlin Zhan, Yangxin Wu, Xiao Dong, Yunchao Wei, Minlong Lu, Yichi Zhang, Hang
  Xu, and Xiaodan Liang.
\newblock Product1m: Towards weakly supervised instance-level product retrieval
  via cross-modal pretraining.
\newblock {\em CoRR}, abs/2107.14572, 2021.

\bibitem{Youcook2}
Luowei Zhou, Chenliang Xu, and Jason~J. Corso.
\newblock Towards automatic learning of procedures from web instructional
  videos.
\newblock In {\em AAAI}, pages 7590--7598, 2018.

\bibitem{singlmodal2}
Tao Zhou, Mingxia Liu, Huazhu Fu, Jun Wang, Jianbing Shen, Ling Shao, and
  Dinggang Shen.
\newblock Deep multi-modal latent representation learning for automated
  dementia diagnosis.
\newblock In {\em MICCAI}, pages 629--638, 2019.

\bibitem{Kaleido-BERT}
Mingchen Zhuge, Dehong Gao, Deng{-}Ping Fan, Linbo Jin, Ben Chen, Haoming Zhou,
  Minghui Qiu, and Ling Shao.
\newblock Kaleido-bert: Vision-language pre-training on fashion domain.
\newblock In {\em CVPR}, pages 12647--12657, 2021.

\end{thebibliography}
}

\clearpage
\setcounter{section}{0}
\renewcommand\thesection{\Alph{section}}

\section{Dataset License}
\label{sec:license}
Our \textbf{M5}Product dataset is released under CC BY-NC-SA 4.0 license and can freely be used for non-commercial purposes. More detailed information can be found at \textcolor[RGB]{254,67,101}{ \url{https://xiaodongsuper.github.io/M5Product_dataset/terms_of_use.html}}, which also provides dataset details and usage guidance. \textbf{Note: For anonymity reasons, the link is not included during the review process.}

\section{Annotation Collection}
\label{sec:dataset_collect}
\begin{figure*}[htbp]
\centering
\includegraphics[width=1\textwidth]{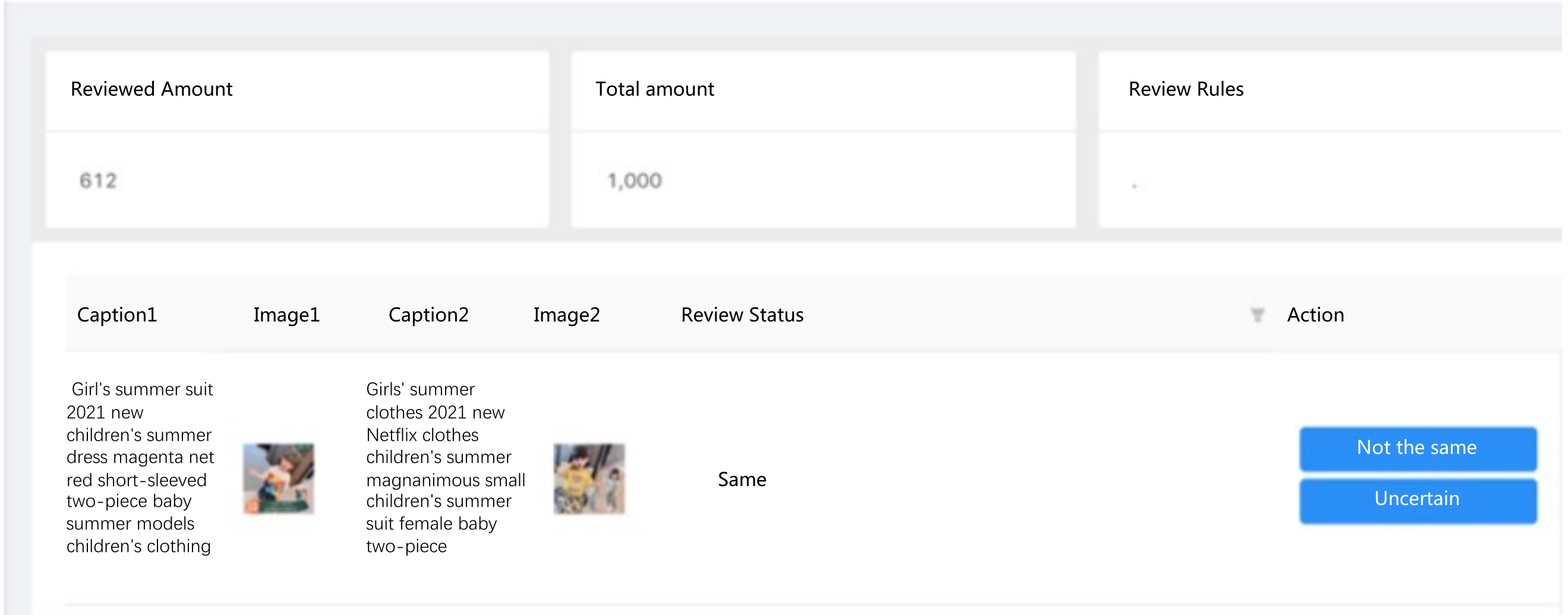}
\caption{UI for huamn annotation on product retrieval task.}
\label{fig:human}
\end{figure*}

We resort to crowd-sourcing to obtain human annotations for the product retrieval task. Specifically, we present human annotators with a matching task, where annotators are asked to select the matching image-text pairs for a given query image-text pair. In our crowdsourcing system, each matching task is presented to five different human annotators and a typical example of our interface is shown in Figure~\ref{fig:human}. The left part of the interface shows the current query data (image and text), while the right side depicts an example from the candidate list. The annotators are then asked to choose from two options: \emph{mismatched} and \emph{uncertain}.
The default labelling option is \emph{matched}.
The interface also displays the number of examples that have been reviewed and the total amount of examples to review.
Each annotation task, can be considered as a binary classification task for the human worker, where he/she has to decide if the pair is a match or not. For each estimated task, the annotators receive a payment of 3 cents RMB.

\section{Annotation}
\label{sec:dataset_anno}
The retrieval task annotation for any query sample consists of all the matched instances in the gallery split. To construct a reliable gallery set, we first use a ResNet50~\cite{Resnet}  and  Bert-Base~\cite{Bert} to extract features and construct the query candidate pool from all the data that is not contained in the training subset.
Specifically, we sample an instance from a category that contains more than 2,000 instances and extract the image and text features. We then concatenate the features and compute the cosine similarity to all other instances of the dataset to produce a pre-ranked candidate list in order to minimize the labelling cost.
The final size of the candidate shortlist for each query is 500, which is about $0.01\%$ of the whole gallery split.  During the crowd-sourced annotation process, human workers review both images and captions in the candidate list to select which samples are matched with the query instance.

\textbf{Annotation Rules.}
It is quite challenging to define whether two images contain the same product when critical aspects are not given in their captions and images. In our annotations, we use product images and their captions as the primary materials for gallery construction. Hence, we define several rules to determine the "same product" condition and provide them as instructions to the annotators. Images contain the same product, if:
\begin{enumerate}[]
\item The two images are in different conditions (e.g., backgrounds, angles, etc), but the products in both images are the same.
\item They should have the same color/model/shape/style, or other features that can be distinguished by humans.
\item The caption has the same product name but the product description differs.
\item They share more than one characteristic such as appearances, materials, colors and so on.
\end{enumerate}

To ensure labeling consistency, each annotation pair is labeled by five human workers in the crowd-sourced platform. In the process, we first make a small dataset from our query list as a Gold Problem to evaluate the annotation capability of each human worker. Based on the labeled results ("Matched" or "Not Matched") from human workers and their annotation capability, we utilize the weighted GLAD~\cite{GLAD} inference algorithm to determine the final accepted labels.

\section{Dataset Split}
\label{sec:dataset_split}

The \textbf{M}5Product dataset is split into several parts to ensure consistent training and evaluation of the models on the various tasks. The \textit{training} set contains 4,423,160 samples from 3,593 classes.

\paragraph{Retrieval} To evaluate models on the retrieval tasks, the remaining data is split into \textit{gallery-c} and \textit{query-c} sets, which are used for the coarse-grained retrieval task, and \textit{gallery-fg} and \textit{query-fg} sets, which are used for the fine-grained retrieval task. The difference between the two retrieval tasks lies in the granularity of their annotation. In the fine-grained task, only identical products are considered a match (for example, all IPHONE 11 Black), while in the coarse-grained task, category labels are being used to group products from each category (for example, all phones are considered a match).

To construct the fine-grained sets, we extracted all cosmetics categories and, using the abovementioned annotation procedure, finally obtained 1,991 \textit{query-fg} samples and 117,858 \textit{gallery-fg} samples.
The \textit{query-c} and \textit{gallery-c} sets contain 24,410 and 1,197,905 samples, respectively. Among the samples in the \textit{gallery-c} set, 249,614 samples are matched with samples in the \textit{query-c} set, while 948,291 samples do not match. These unmatched samples are added to the \textit{gallery-c} set to increase the difficulty of the retrieval task.
We further report the finetuned retrieval performance in the paper, which corresponds to retrieval performance after finetuning the model using the classification training set (see next paragaraph).


\paragraph{Finetuning, Classification and Clustering}


For the classification and clustering tasks, we sampled 1,805 categories from the whole dataset and obtained 18,526 train samples and 4,632 test samples. We first finetune our \textbf{SCALE} using the classification training set and then extract features from the finetuned model to perform the classification and clustering tasks.

\section{Data Format}
\label{sec:dataset_dataformat}


The dataset consists of 6,313,067 products uploaded by 1,000,517 merchants, where merchant information has been removed to ensure anonymity. In the following, we outline the different modalities:

\noindent\textbf{Image data} Each product has at least five product images, where the first image is the main image that gives the detailed overview of the product, while the rest depict its functionalities or characteristics. We pick all the main images to construct the dataset.


\noindent\textbf{Caption/text data} are provided by the 1,000,517 merchants. Note that the text description does not always match well with the other modalities.

\noindent\textbf{Video data} are used to showcase the products' usage and characteristics to customers. In our dataset, these videos are recorded at a speed of 24 frames per second (FPS). To reduce the amount of redundant information that is contained in adjacent frames and the dataset as a whole, we only select one frame per second.

\noindent\textbf{Audio data} are extracted from the video data. We extract the corresponding audio information of all sampled video frames. Then the audio frames are transformed into spectrograms using Mel-Frequency Cepstral Coefficients (MFCC)\cite{2005Combining}. We set the frame size and hop size as 1,024 and 256, respectively.

\noindent\textbf{Tabular data} are a special kind of database that records some additional product characteristics such as appearance, purpose and producer. The tabular data is indexed by the product ID and collected from the whole product database. There are 5,679 different types of property information and 24,398,673 unique values.


\begin{table*}
\centering
\caption{Comparisons with other widely used multi-modal datasets. "-" means not mentioned.  Our \textbf{M}5Product is one of the largest  multi-modal datasets compared with existing datasets.}
\label{tab:dataset_compare_all}
\setlength{\tabcolsep}{0.4mm}
\resizebox{2\columnwidth}{!}{
\begin{tabular}{c|cccc|ccc|c}
\toprule[1pt]
{{Dataset}}& {\footnotesize{}{Samples}} &  {\footnotesize{}{Categories}} &  {\footnotesize{}{Instances}}
&  {\footnotesize{}{Modalities}}
&  {\footnotesize{}{Modal type}}  & {\footnotesize{}{Product}}
\tabularnewline
\midrule[1pt]
LJ Speech~\cite{ljspeech17} & 13,100 & - & - & 2 &audio/text &no  \tabularnewline
SQuAD~\cite{SquAD} & 37,111  & - & - & 2 &audio/text &no  \tabularnewline
TVQA~\cite{TVQA} & 21,793  & - & - & 2 &video/text &no  \tabularnewline
MovieQA~\cite{MovieQA} & 408  & - & - & 2 &video/text &no  \tabularnewline
TGIF-QA~\cite{TGIF} & 56,720  & - & - & 2 &video/text &no  \tabularnewline
AVSD~\cite{AVSD} & 11,816  & - & - & 2 &video/text &no  \tabularnewline
Youcook2~\cite{Youcook2} & 14,000  & 89 & - & 2 &video/text &no  \tabularnewline
VATEX~\cite{VATEX} & 35,000  & - & - & 2 &video/text &no  \tabularnewline
MSRVTT~\cite{MSRVTT} &100,000  & 20 & - & 2 &video/text &no  \tabularnewline
HowTo100M~\cite{HowTo100M} &1,220,000  & 12 & - & 2 &video/text &no  \tabularnewline
Conceptual Caption 3M~\cite{CC} &3,300,000  & - & - & 2 &image/text &no  \tabularnewline
SBU~\cite{SBU} &890,000  & - & - & 2 &image/text &no  \tabularnewline
Visual Genome~\cite{VG} &108,000  & - & - & 2 &image/text &no \tabularnewline
COCO~\cite{COCO} &123,287  & - & - & 2 &image/text &no  \tabularnewline
Flickr30K~\cite{Flickr} &31,000  & - & - & 2 &image/text &no  \tabularnewline
NLVR2~\cite{NLVR2} & 107,292   & - & - & 2 &image/text &no  \tabularnewline
VQA2.0~\cite{VQA} & 204,721   & - & - & 2 &image/text &no  \tabularnewline
RPC checkout~\cite{RPC} & 30,000 & 200 & 367,935 & 2 &image/text &no  \tabularnewline
Twitter100k~\cite{twitter100k} & 100,000 & - & - & 2 & image/text &no   \tabularnewline
INRIA-Websearch~\cite{INRIA} &  71,478 & 353  & - &2 & image/text &no  \tabularnewline
NUS-WIDE~\cite{nuswide} &  269,648 & 81  & - &2 & image/text &no   \tabularnewline
Open Image~\cite{OpenImage} & 1,670,000 & - & - &2 & image/text &no  \tabularnewline
Conceptual 12M~\cite{CC12M} & 12,423,374 & - & - &2 & image/text &no  \tabularnewline
CMU-MOSEI~\cite{cmumosei} &23,500  & 2 & - & 3 &text/video/audio &no \tabularnewline
XMedia~\cite{XMedia} & 12,000  &20  & - &5 & image/text/video/audio/3D &no \tabularnewline
\midrule[1pt]
Dress Retrieval~\cite{dress} & 20,200 & 50 & $\sim$20,200 &2 & image/text &yes \tabularnewline
MEP-3M~\cite{MEP} & 3,012,959 & 599 & - &2 & image/text &yes  \tabularnewline
Product1M~\cite{Product1M} & 1,182,083  & 458 & 92,200 &2 & image/text &yes \tabularnewline
\textbf{M}5Product & \textbf{6,313,067}  & \textbf{6,232} & - &\textbf{5} & \textbf{image/text/video/audio/table} &\textbf{yes} \tabularnewline
\bottomrule[1pt]
\end{tabular}}
\end{table*}

\begin{table*}[t!]
\centering
\vspace{-4mm}
\caption{The retrieval performance with missing modalities.}
\label{tab:modal_missing}
\tiny
\resizebox{2\columnwidth}{!}{
\begin{tabular}{
    c |
    c|
	c
	c
	c|
	c
	c
	c
}
\toprule[1pt]
{Modal}  & {Accuracy} &{mAP@1} & {mAP@5} & {mAP@10}  & {Prec@1} & {Prec@5} & {Prec@10}\\
\midrule
\textbf{SCALE} (full-modality) & 84.06 & 57.97/ 69.12 & 62.54/ 71.93 & 60.48 / 69.92 & 57.97 / 69.12 & 38.02 / 47.63 & 28.88 / 34.70 \\
\textbf{SCALE} & 85.50 & 58.72 / 70.62 & 63.17 / 73.02 & 61.05 / 71.50 &  58.72 / 70.62 & 39.66 / 48.20 & 30.32 / 35.35 \\
\bottomrule[1pt]
\end{tabular}}
\vspace{-4mm}
\end{table*}

\begin{figure*}[!htp]
\centering
\includegraphics[width=0.32\textwidth]{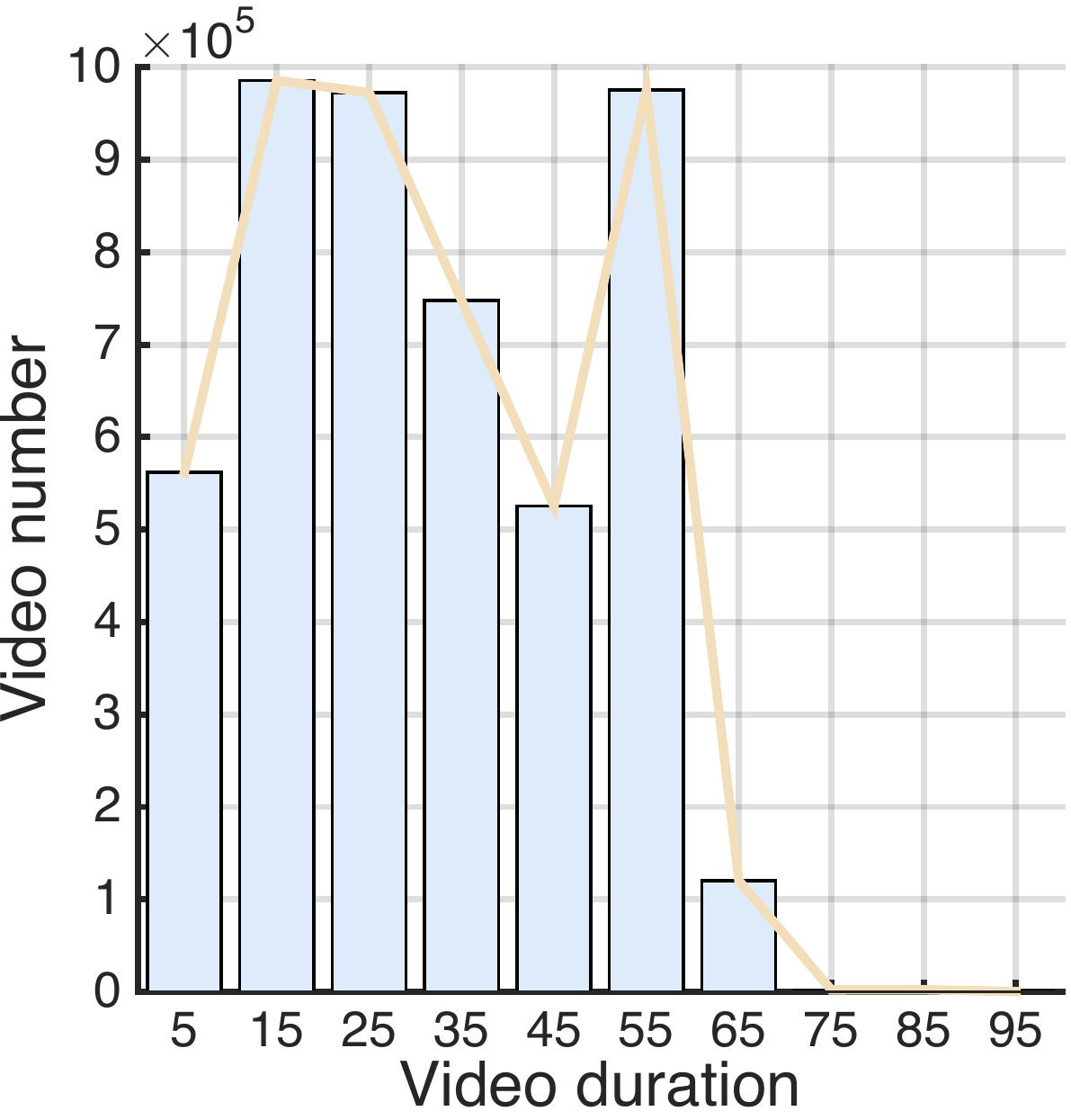}
\includegraphics[width=0.32\textwidth]{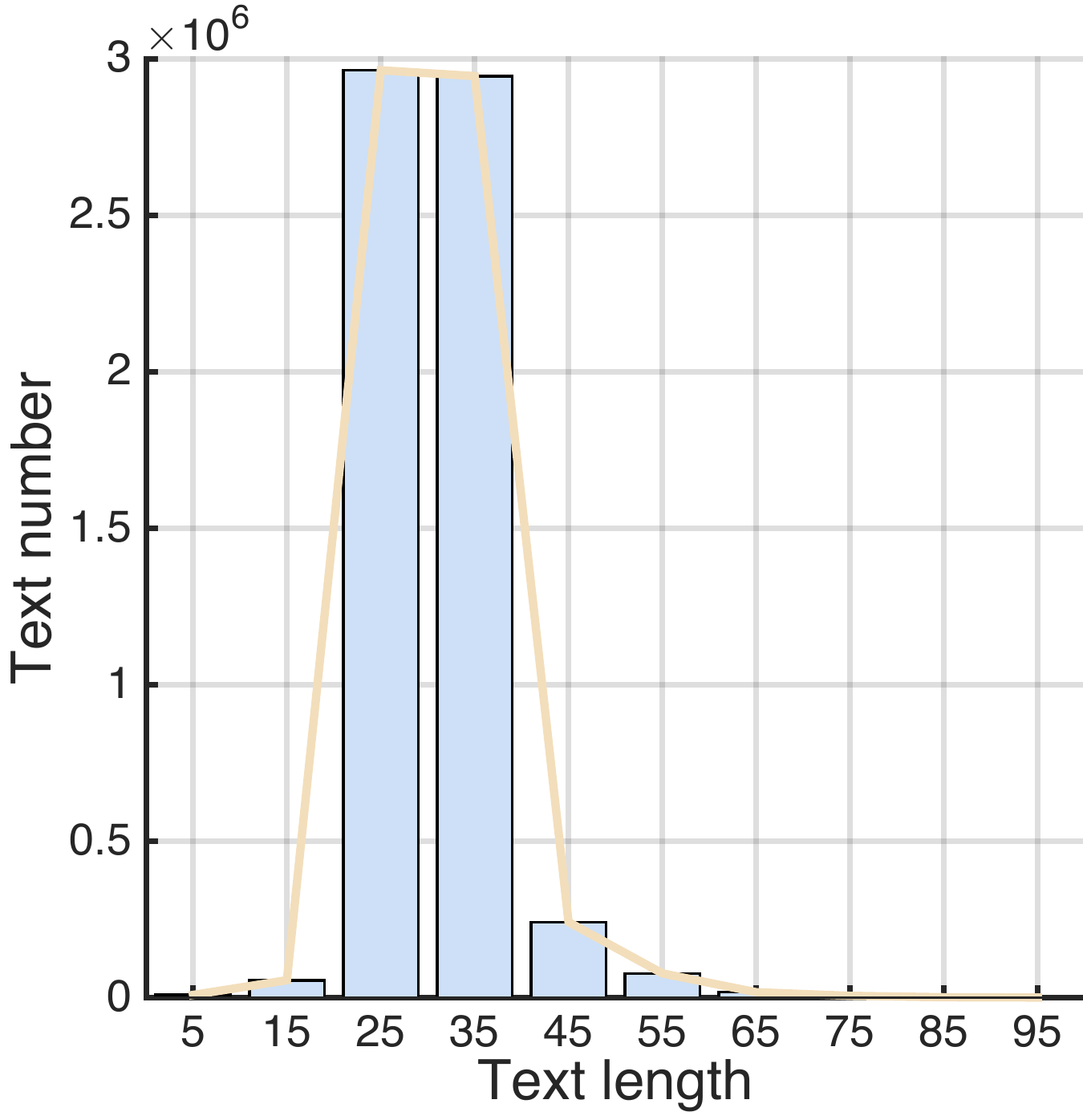}
\includegraphics[width=0.315\textwidth]{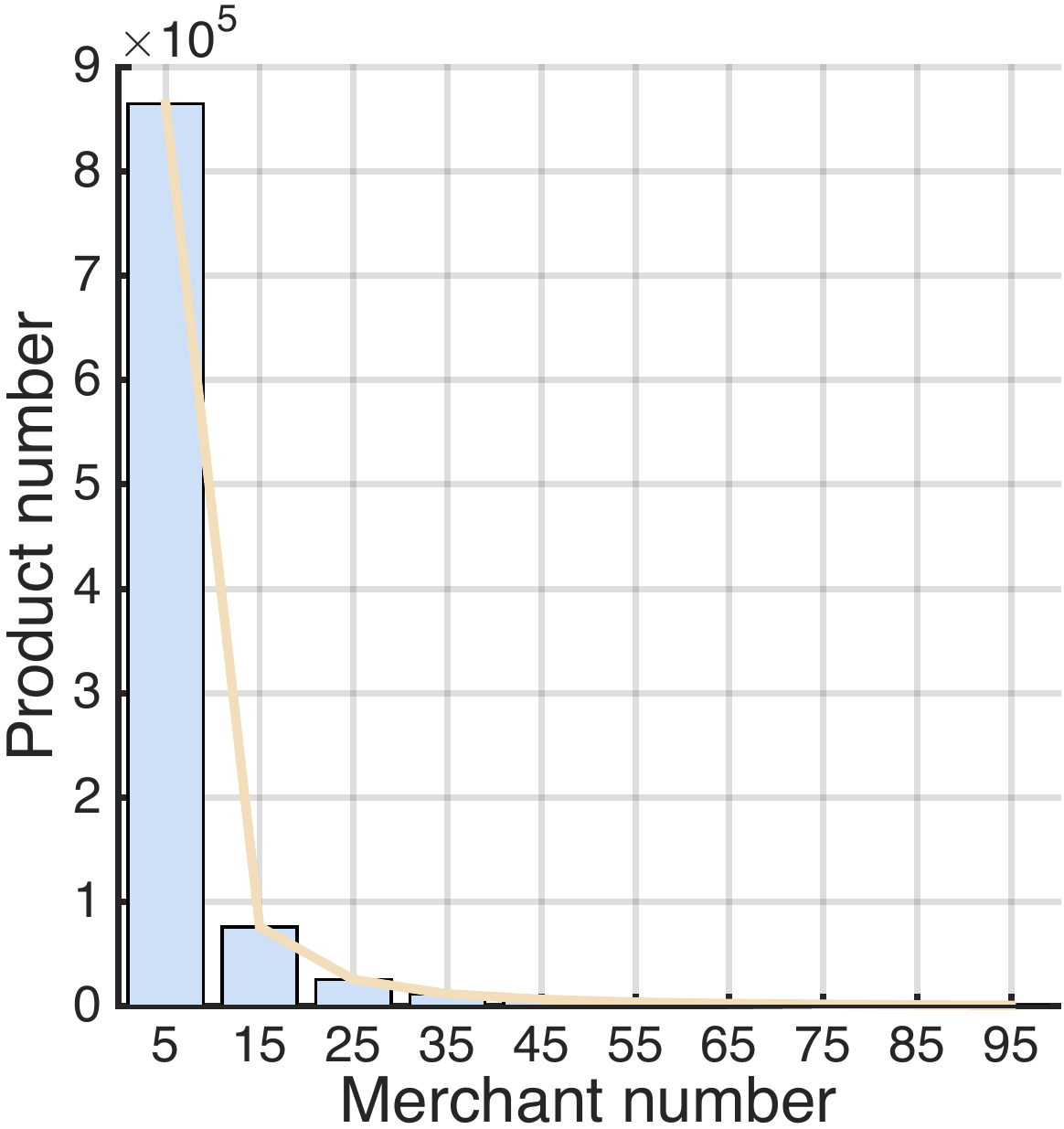}
\caption{The distributions of video, text and merchant on our \textbf{M5}Product.}
\label{fig:data_modality_distribution}
\end{figure*}

\section{Unimodal and unpair analysis}
\label{sec:dataset_dataanalysis}
1) \textbf{Unimodal analysis}: Figure~\ref{fig:data_modality_distribution} gives the video, text and merchant distributions. From the figure, we can find that the video duration, the text length and the merchant number range from 1 to 60 seconds, 20 to 40 words and 1 to 10 product numbers, respectively. This variation further illustrates the real-world nature of our dataset.
2) \textbf{Unpair analysis}: In the data collection process, there are 82,577 invalid URLs for the image modalities (1.3$\%$ of the products), while the number of samples that contain both the Image and Text modalities is 6,230,490.
Further taking into account the table modality, the number of complete samples drops by 1.4$\%$ to 6,225,598 samples that have all three modalities.
Overall, the dataset contains 5,050,078 samples that contain all five modalities. This means that about 20$\%$ of the samples are incomplete. This is mostly due to merchants being biased towards specific modalities, which is a common scenario in the real world.

\section{Implementation Details}
\label{sec:implement}
Our models are implemented in Pytorch~\cite{pytorch}. To speed up training, we use Nvidia Apex\footnote{https://github.com/NVIDIA/apex} for mixed precision training. All models are trained on 4 Nvidia 3090 and 2080ti GPUs on our workstations. We use Adam~\cite{Adam} to optimize the parameters of our model, with an initial learning rate of 1\textbf{e}-4, and use a linear learning rate decay schedule with a temperature parameter of 0.1.

\section{Dataset Comparison}
\label{sec:dataset_comp}
A comprehensive comparison between our \textbf{M5}Product dataset and other widely used multi-modal pre-training datasets is shown in Table~\ref{tab:dataset_compare_all}. From the table, we can observe that our \textbf{M5}Product not only has more diverse modalities but also contains a large amount of data samples from an abundant amount of categories.

\section{Missing data verification}
\label{sec:missing_data}
Results in Table~\ref{tab:modal_missing} show the superiority of our methods over the standard approach of ignoring incomplete samples. We compare two variants of our \textbf{SCALE} framework: 1) \textbf{SCALE} (full-modality) and our proposed \textbf{SCALE}. The only difference between the two methods is the input. The input of the former only includes complete samples (all modalities present), while the input of the latter includes the incomplete modality samples. The verification is performed on the subset dataset as mentioned in the main article.

\section{Failure Analysis}
\label{sec:failure}
\begin{figure*}[!htp]
\centering
\includegraphics[width=0.95\textwidth]{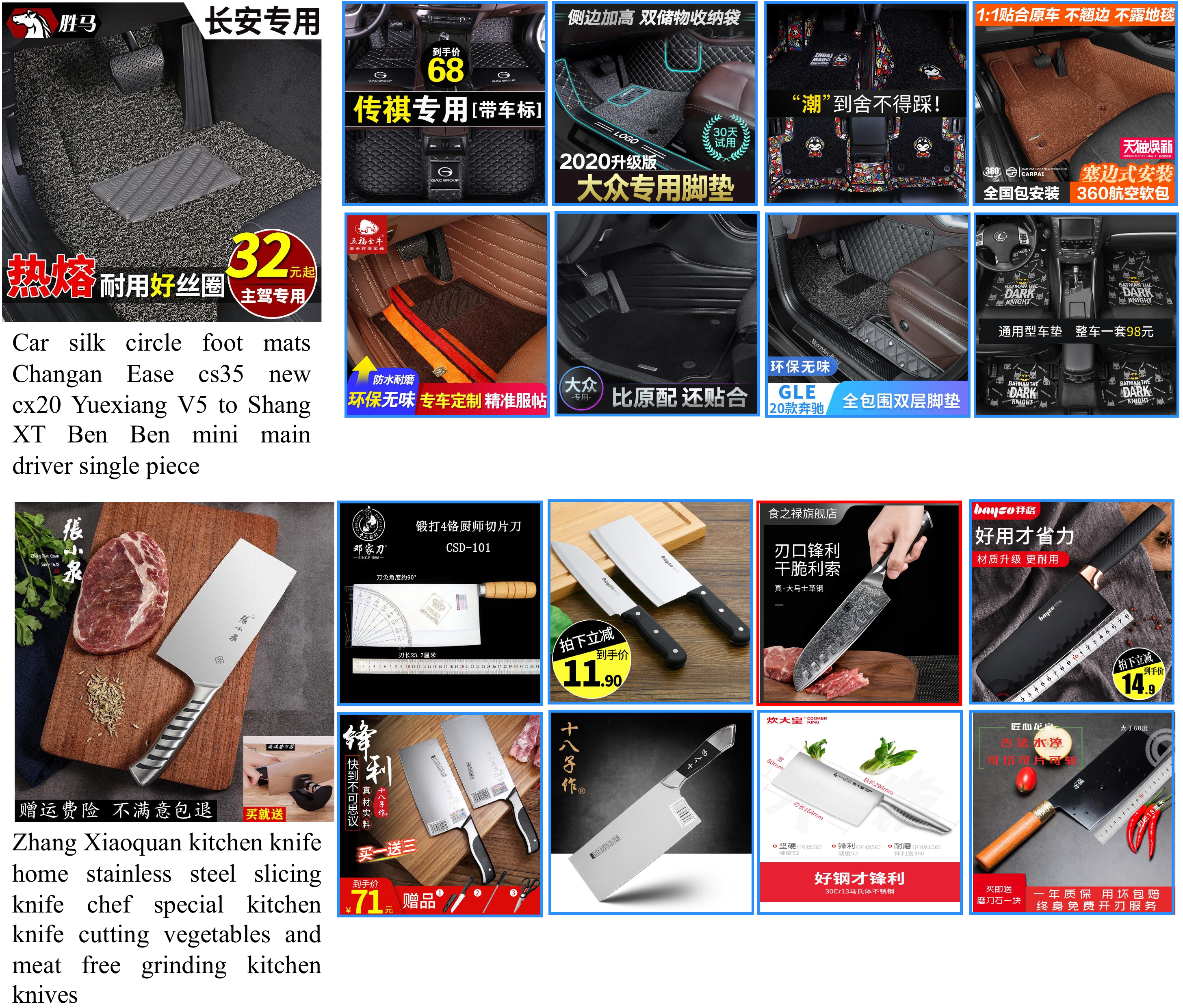}
\caption{Successful retrieval results 1 by our \textbf{SCALE}.}
\label{fig:failure_3}
\end{figure*}

\begin{figure*}[!htp]
\centering
\includegraphics[width=0.95\textwidth]{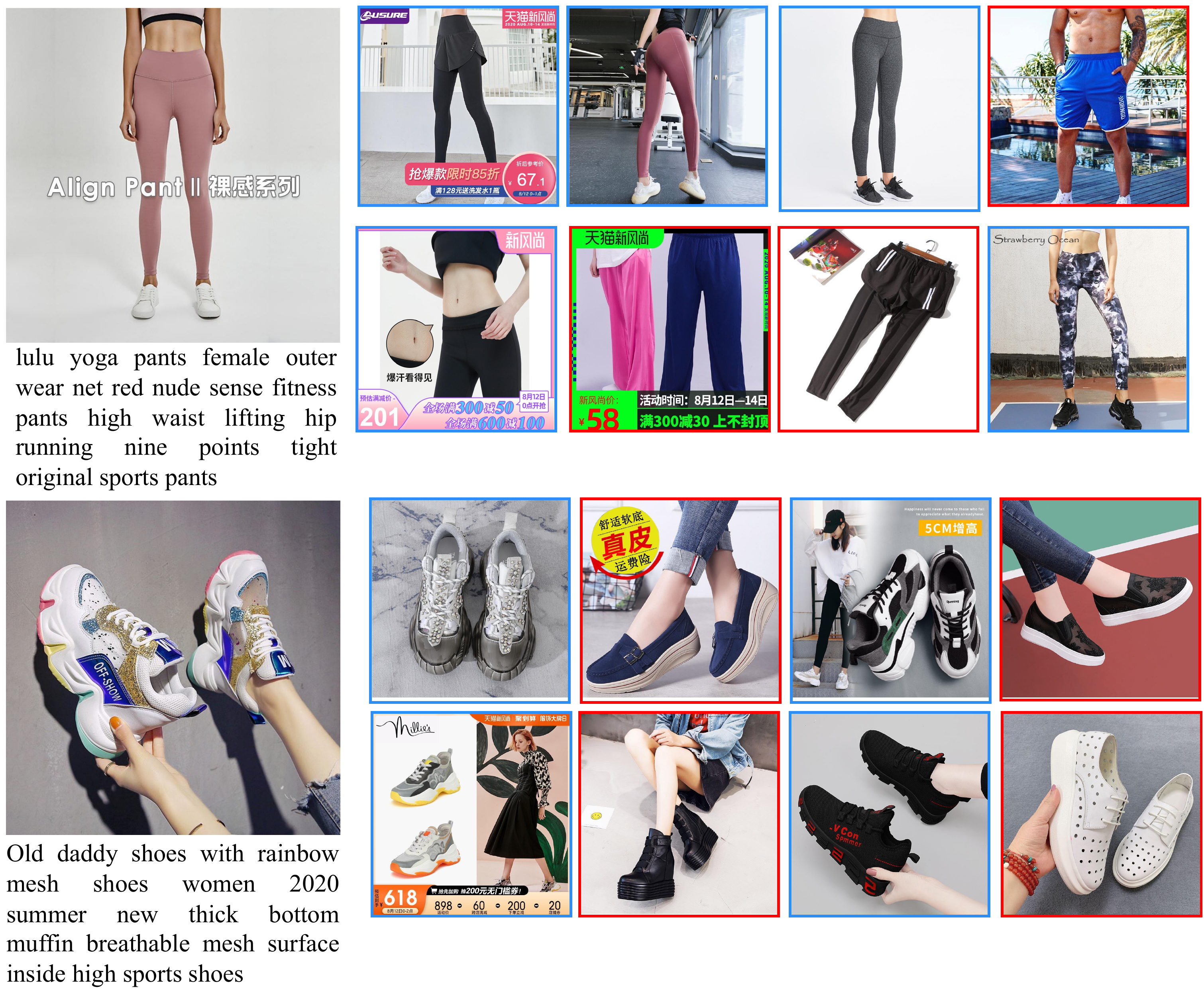}
\caption{Failure retrieval results 2 by our \textbf{SCALE}.}
\label{fig:failure_1}
\end{figure*}

\begin{figure*}[!htp]
\centering
\includegraphics[width=0.95\textwidth]{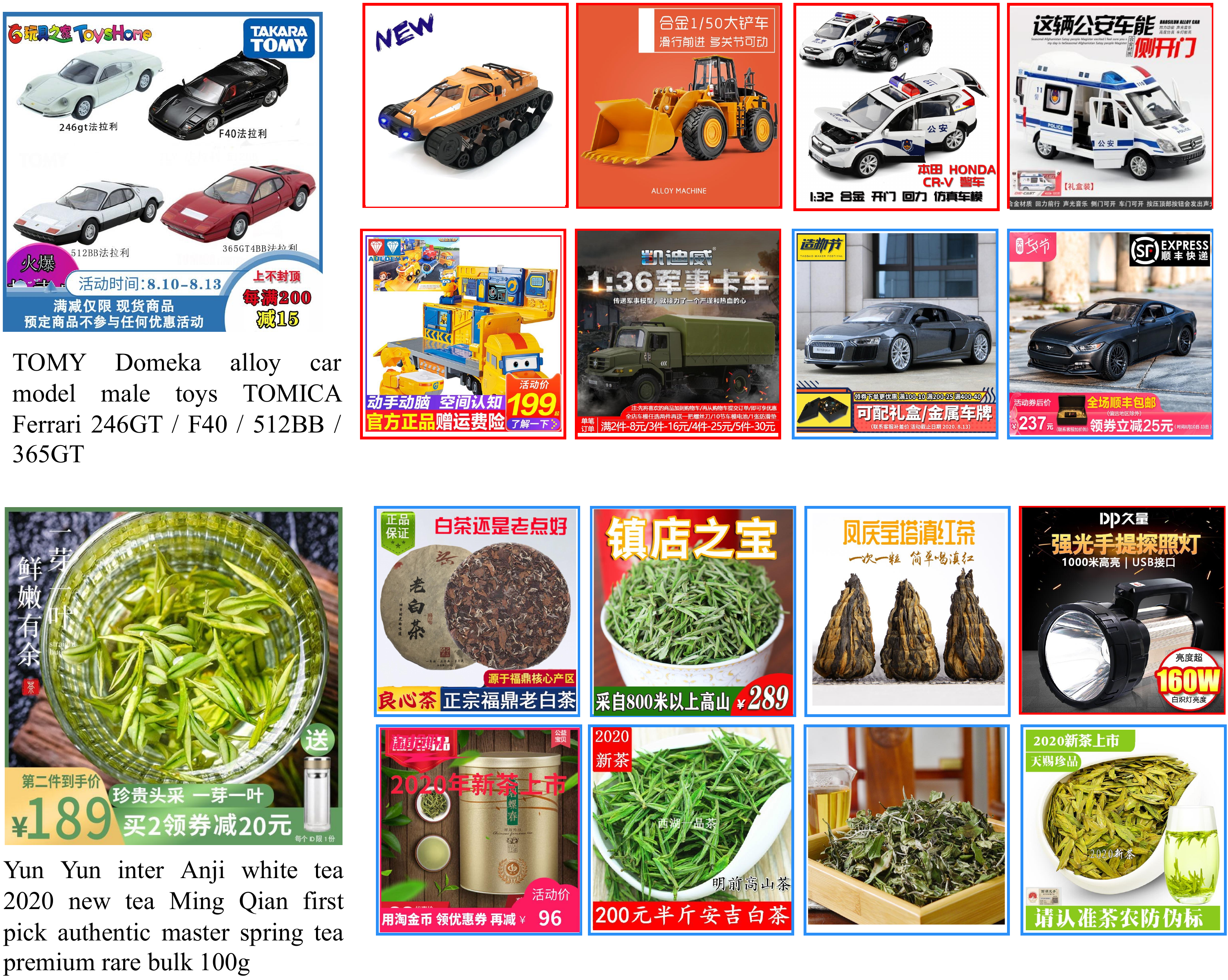}
\caption{Failure retrieval results 3 by our \textbf{SCALE}.}
\label{fig:failure_2}
\end{figure*}

Several product retrieval examples are shown in Figures~\ref{fig:failure_3}, \ref{fig:failure_1}, and \ref{fig:failure_2}. The first column represents the image and text modality of the query sample, while the eight images to its right belong to the matched results from the gallery set. In the matched results, the samples boxed in blue are the correctly matched samples, while the samples boxed in red are mismatched. These retrieval results illustrate that the learned embeddings are discriminative. However, in a few cases, the recalled samples are not matched due to the limited number of category samples in the gallery set or similar descriptions in the text data.

\section{More Visualization}
\label{sec:visuliazation_more}

\begin{figure*}[!htp]
\centering
\includegraphics[width=0.95\textwidth]{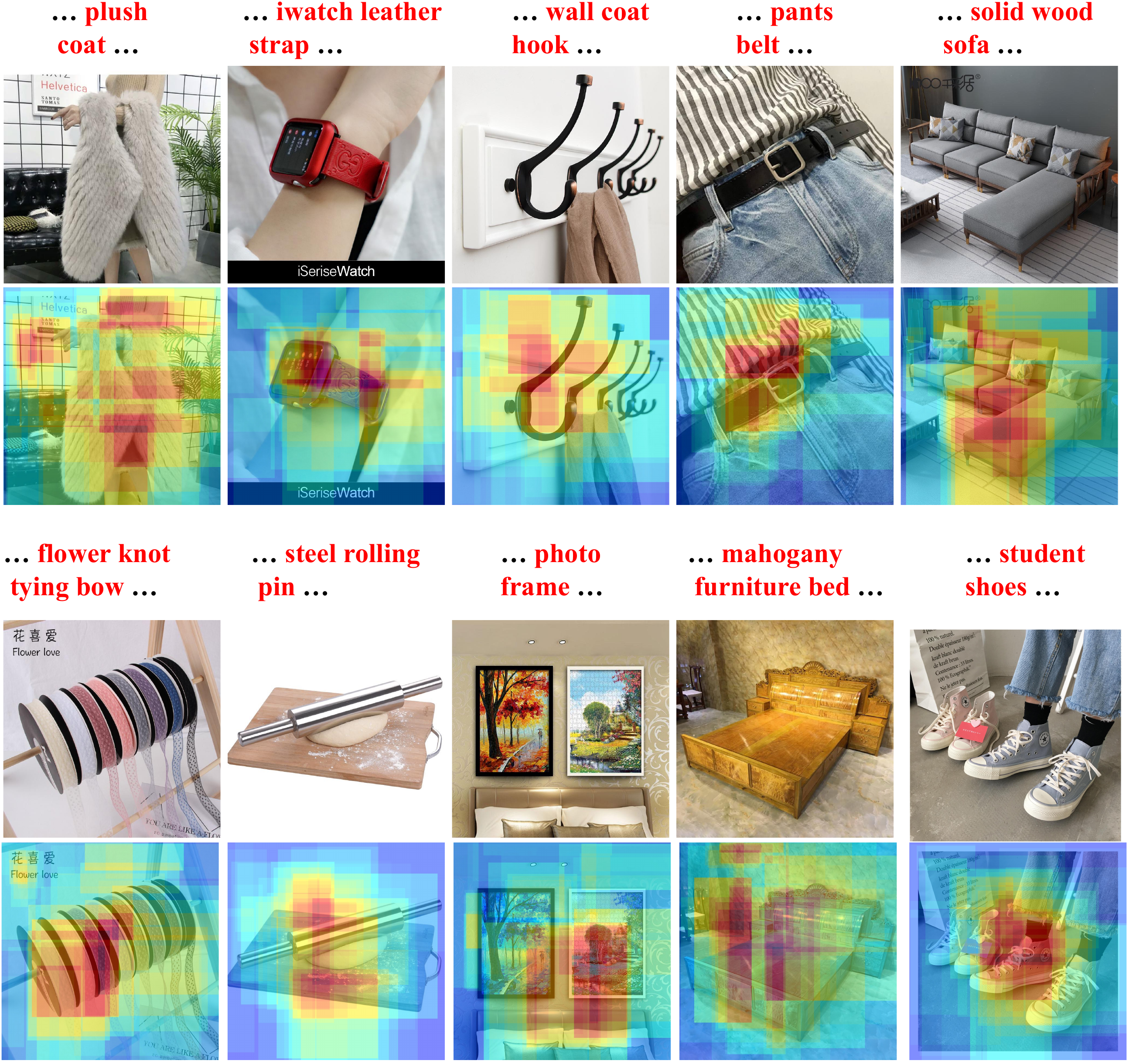}
\caption{More attention visualization 1 by our \textbf{SCALE}.}
\label{fig:attention_more_1}
\end{figure*}

\begin{figure*}[!htp]
\centering
\includegraphics[width=0.95\textwidth]{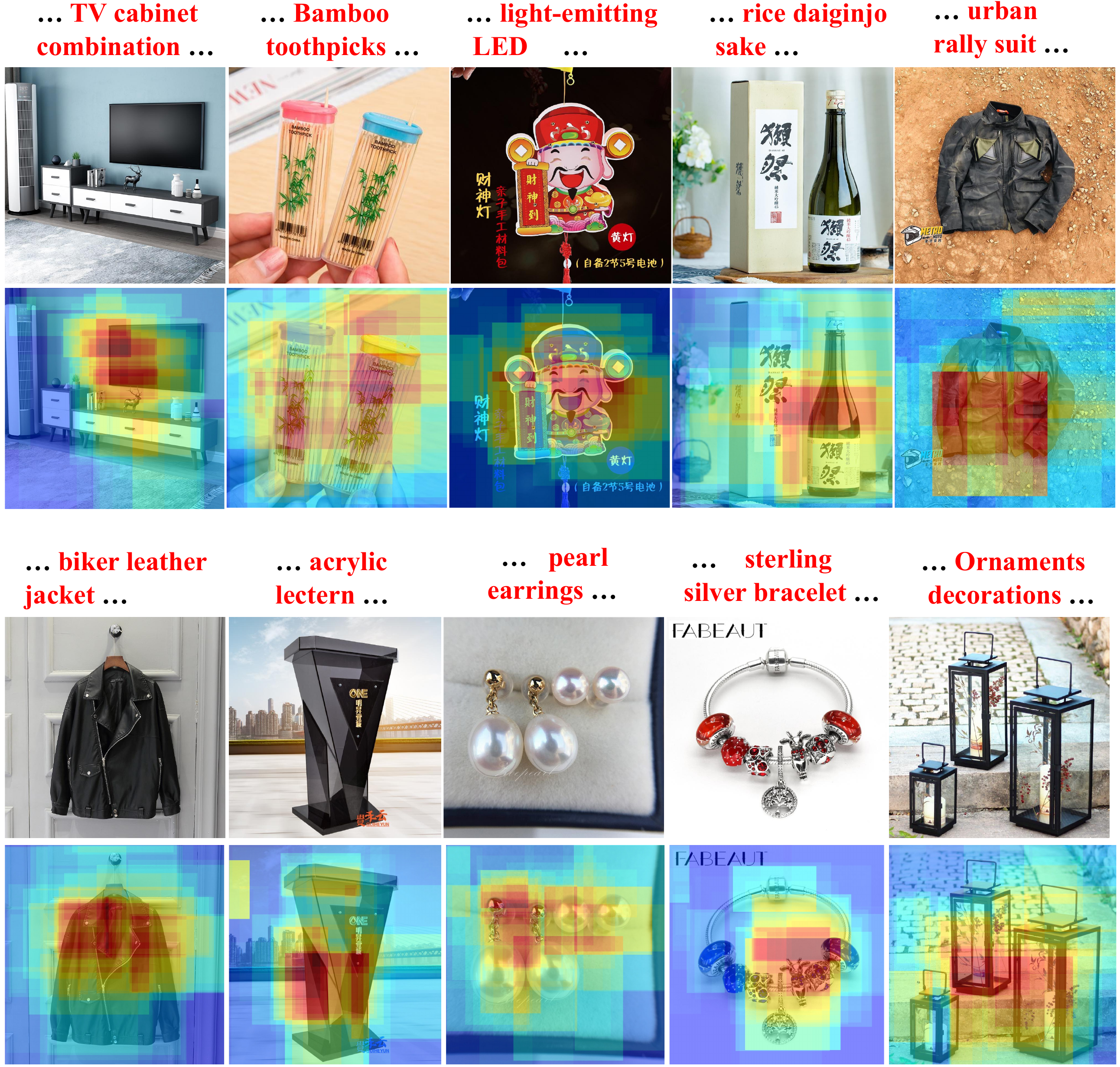}
\caption{More attention visualization 2 by our \textbf{SCALE}.}
\label{fig:attention_more_2}
\end{figure*}

Additional attention visualization results are provided in Figures~\ref{fig:attention_more_1} and \ref{fig:attention_more_2}. Similar to the illustrations in the main paper, these visualizations show that \textbf{SCALE} can learn the detailed semantics in the images and the text.

\section{Code and Our dataset.}
The code is provided in the supplementary and we also include a few full-modality examples of the dataset. Due to space constraints, we have refrained from sharing the whole dataset.

\end{document}